\documentclass[lettersize,journal]{IEEEtran}
\usepackage{amsmath,amsfonts}
\usepackage{algorithmic}
\usepackage{algorithm}
\usepackage{array}
\usepackage[caption=false,font=normalsize,labelfont=sf,textfont=sf]{subfig}
\usepackage{textcomp}
\usepackage{stfloats}
\usepackage{url}
\usepackage{float}

\usepackage{rotating}
\usepackage{tablefootnote}
\usepackage{multirow}
\usepackage{paralist}
\usepackage{booktabs}
\usepackage[table]{xcolor}
\usepackage{marvosym}
\usepackage{pifont}
\usepackage{amssymb}
\usepackage[normalem]{ulem}
\usepackage{colortbl}
\definecolor{hl}{RGB}{240,240,240}
\definecolor{hler}{RGB}{230,230,200}
\usepackage{verbatim}
\usepackage{graphicx}
\usepackage{cite}
\usepackage{xcolor}

\def\blue#1{\textcolor{blue}{#1}}

\definecolor{darkgreen}{RGB}{0,100,0}
\def\eg{\textit{e.g.}}
\def\ie{\textit{i.e.}}

\usepackage[colorlinks=true,urlcolor=magenta,citecolor=blue,breaklinks=true,bookmarks=false]{hyperref}
\begin{document}

\title{Celeb-DF++: A Large-scale Challenging Video DeepFake Benchmark for Generalizable Forensics}

\author{Yuezun Li, Delong Zhu, Xinjie Cui, Siwei Lyu,~\IEEEmembership{Fellow,~IEEE}
\thanks{Yuezun Li is corresponding author.}
\thanks{Yuezun Li, Delong Zhu, and Xinjie Cui are with the School of Computer Science and Technology, Ocean University of China, Qingdao, China. Email: liyuezun@ouc.edu.cn;\{zhudelong;cuixinjie\}@stu.ouc.edu.cn.}
\thanks{Siwei Lyu is with the University at Buffalo, SUNY, USA. Email: siweilyu@buffalo.edu.}}

\markboth{Journal of \LaTeX\ Class Files,~Vol.~14, No.~8, August~2021}%
{Shell \MakeLowercase{\textit{et al.}}: A Sample Article Using IEEEtran.cls for IEEE Journals}


\maketitle

\begin{abstract}
The rapid advancement of AI technologies has significantly increased the diversity of DeepFake videos circulating online, posing a pressing challenge for \textit{generalizable forensics}, \ie, detecting a wide range of unseen DeepFake types using a single model. 
Addressing this challenge requires datasets that are not only large-scale but also rich in forgery diversity. However, most existing datasets, despite their scale, include only a limited variety of forgery types, making them insufficient for developing generalizable detection methods. Therefore, we build upon our earlier Celeb-DF dataset and introduce {Celeb-DF++}, a new large-scale and challenging video DeepFake benchmark dedicated to the generalizable forensics challenge. Celeb-DF++ covers three commonly encountered forgery scenarios: Face-swap (FS), Face-reenactment (FR), and Talking-face (TF). Each scenario contains a substantial number of high-quality forged videos, generated using a total of 22 various recent DeepFake methods. These methods differ in terms of architectures, generation pipelines, and targeted facial regions, covering the most prevalent DeepFake cases witnessed in the wild. We also introduce evaluation protocols for measuring the generalizability of 24 recent detection methods, highlighting the limitations of existing detection methods and the difficulty of our new dataset. 
This dataset has been released at \url{https://github.com/OUC-VAS/Celeb-DF-PP}.

\end{abstract}


\section{Introduction}
\IEEEPARstart{D}{eepFake} refers to AI-based face synthesis techniques that can easily create falsified videos with high realism. Given the strong association between faces and identity, well-crafted DeepFakes can create illusions of individuals engaging in events that never occurred, resulting in severe political, social, financial, and legal concerns~\cite{folorunsho2025deepfake, busacca2023deepfake, tahraoui2023defending}.

To counter the DeepFakes, numerous detection methods have been developed in recent years, \eg,~ \cite{afchar2018mesonet,cao2022end,li2018ictu,yang2018exposing,matern2019exploiting,li2019exposing,nguyen2019capsule,qian2020thinking,sun2022information,yan2023ucf,yan2024effort,cui2024forensics, qi2020deeprhythm}, necessitating the need of large-scale datasets for training and evaluation. To date, many datasets have been released, such as UADFV~\cite{yang2018exposing}, DeepFake-TIMIT~\cite{korshunov2018deepfakes}, FaceForenscics++~\cite{rossler2019faceforensics++}, DFD~\cite{DDD_GoogleJigSaw2019}, DFDC~\cite{dolhansky2020deepfake}, Celeb-DF~\cite{li2020celeb}, etc. While these datasets have significantly accelerated progress in DeepFake detection,
they are built using early DeepFake methods, thus often lacking diversity in forgery types.

\begin{figure}[!t]
    \centering
    \includegraphics[width=\linewidth]{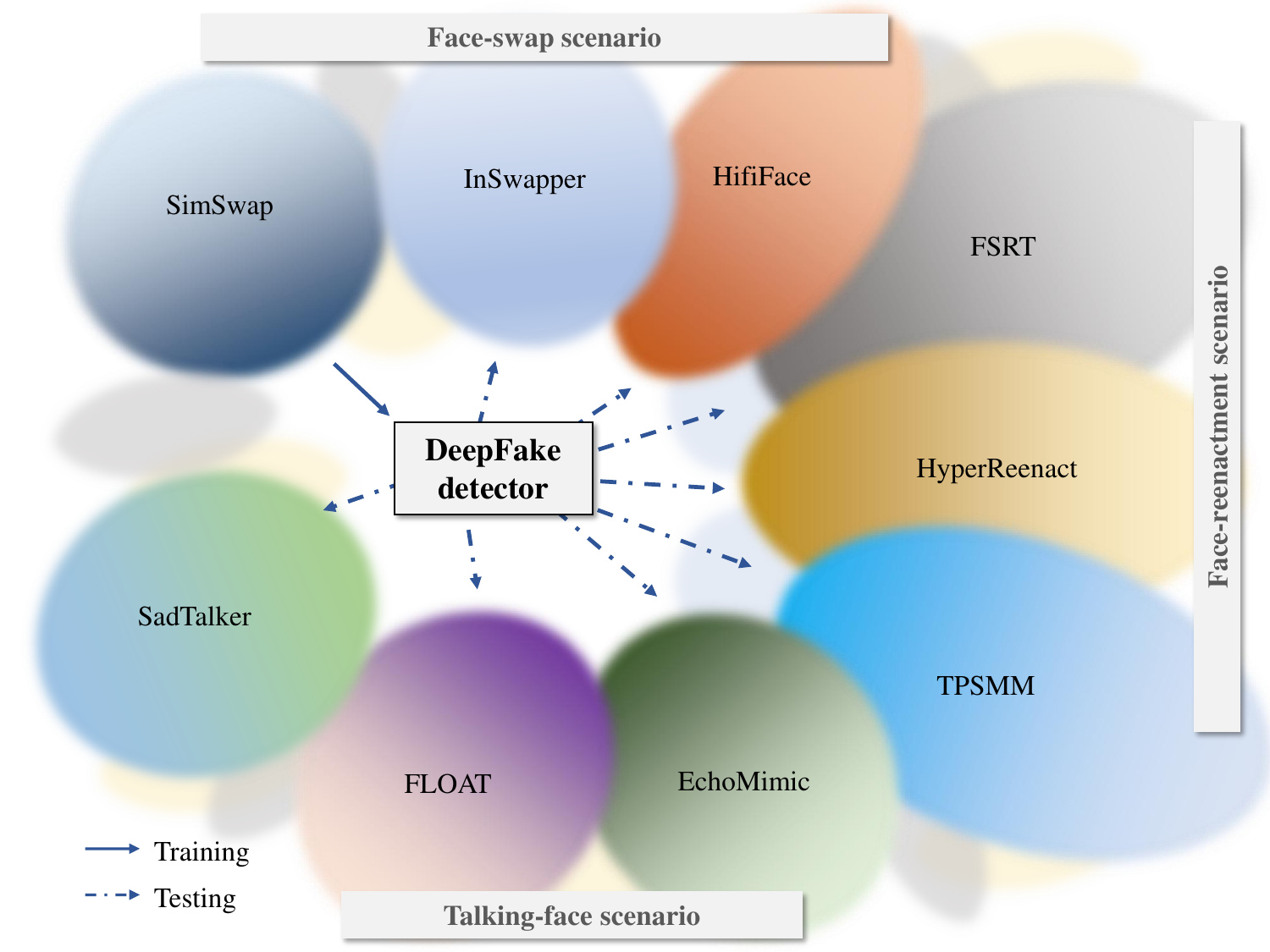}
    \caption{\small Overview of the proposed Celeb-DF++ benchmark. This benchmark is motivated by the need of \textit{generalizable forensics} in deepfake detection.}
    \label{fig:overview}
\end{figure}

In recent years, the rapid advancement of generative models (\eg, GAN~\cite{goodfellow2014generative}, Diffusion model~\cite{ho2020denoising}) has led to a surge in the developement of DeepFake methods, greatly increasing their diversity. With continuous architectural refinements and evolving generation strategies, DeepFakes become increasingly diverse, and more concerningly, many of their details are likely unknown. This circumstance raises a practical and urgent challenge for DeepFake detection, referred to as \textit{Generalizable forensics: Can a detector effectively identify a wide range of unseen DeepFakes?} (see Fig.~\ref{fig:overview}) 
However, due to the limited diversity of forgeries in existing datasets, they fall short in supporting the development of detection methods that generalize well in the wild. This highlights the critical need for constructing a large-scale and diverse dataset to drive progress in DeepFake forensics. 

To address this gap, we introduce \textit{Celeb-DF++}, which includes a large variety of DeepFake methods, covering three commonly witnessed scenarios: \textit{Face-swap (FS)}, \textit{Face-reenactment (FR)}, and \textit{Talking-face (TF)}, respectively\footnote{Whole-face synthesis and face editing are also prevalent DeepFake forms. However, as they focus on image manipulation and struggle to maintain temporal consistency across frames, they are rarely applied to videos. Given that videos are generally more deceptive and impactful, we exclude these types from our benchmark.}. In the FS scenario, the face of a source individual is replaced with a synthesized face of a target individual, while retaining consistent facial attributions. FR involves generating new videos of a target individual driven by the behaviors of a source individual, ensuring behavioral consistency. TF leverages audio input to generate synchronized lip movements in synthesized videos of the target individuals. 
For each scenario, we adopt a diverse set of state-of-the-art DeepFake methods, \textbf{8} for FS, \textbf{7} for FR, and \textbf{7} for TF, covering \textbf{22} representative methods in total. These methods vary in architectures, generation pipelines, and manipulated facial regions, effectively simulating the diversity encountered in real-world media.

\begin{figure*}[t]
    \centering
    \begin{minipage}[t]{0.48\textwidth}
        \includegraphics[width=\linewidth]{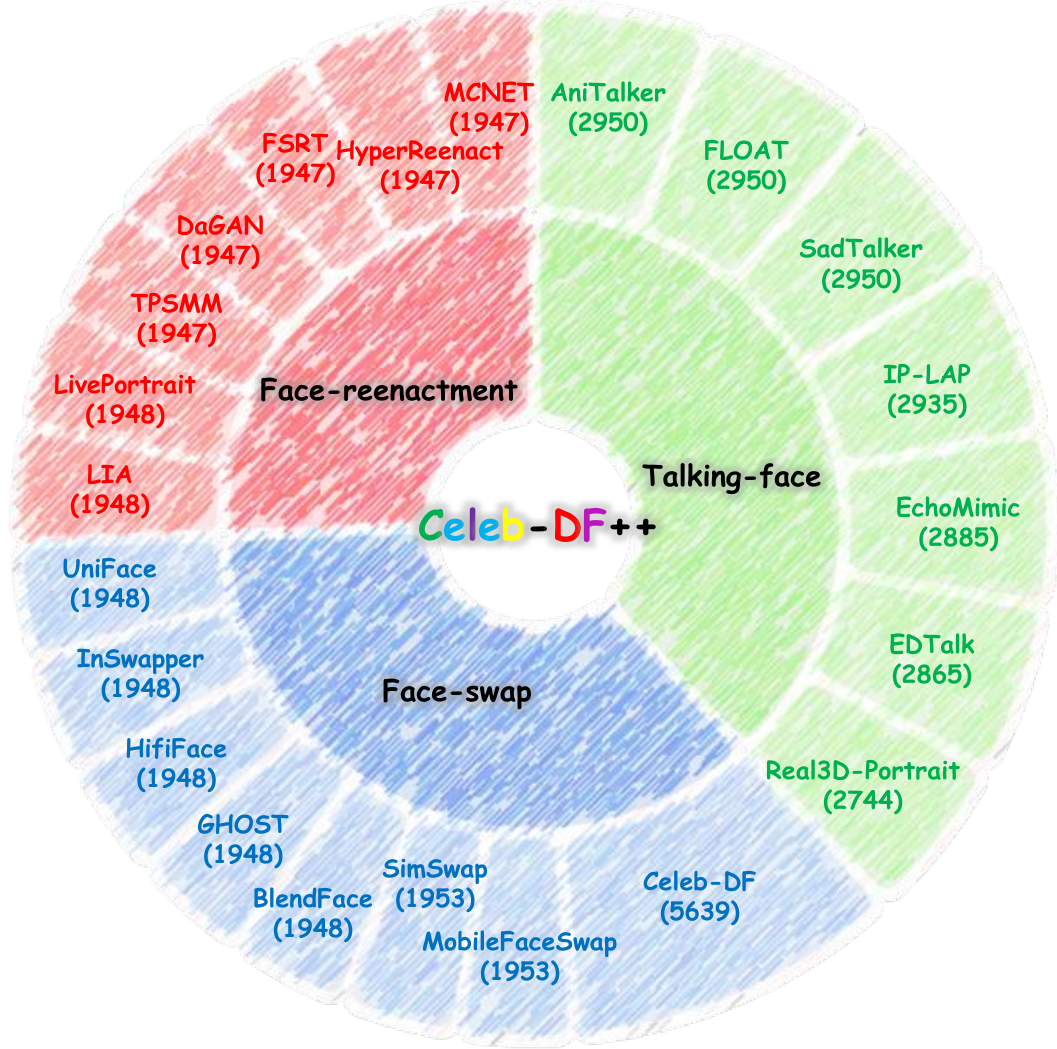}
    \end{minipage}
    \hfill
    \begin{minipage}[t]{0.48\textwidth}
        \includegraphics[width=\linewidth]{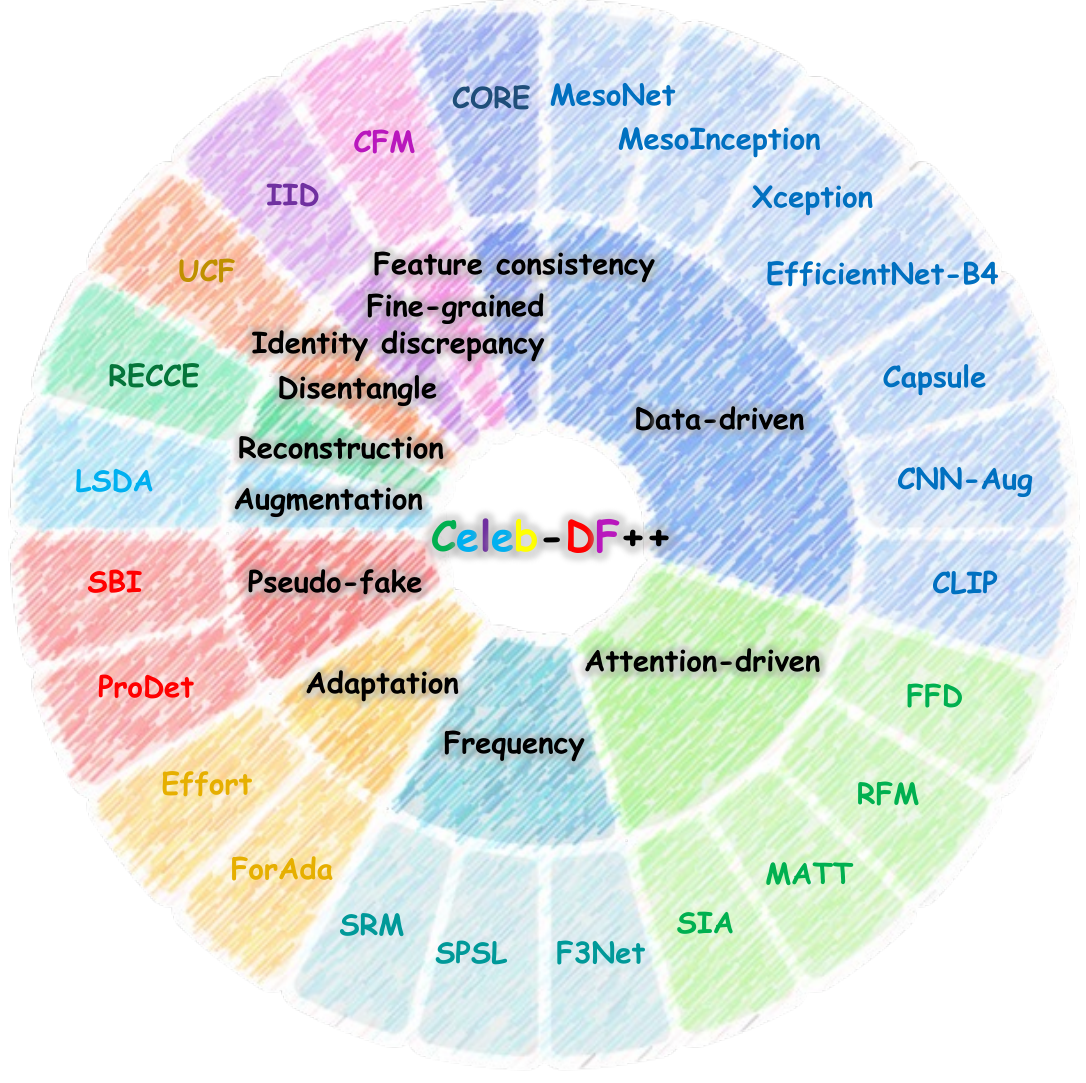}
    \end{minipage}

    \caption{\small {The sunburst chart highlights that our Celeb-DF++ includes a wide range of both DeepFake methods and assessed detection methods. See Table~\ref{table:datasets}, Table~\ref{tab:involved_deepfake} and Table~\ref{tab:detection-methods} for details.}}
    \label{fig:Sunburst}
\end{figure*}

Celeb-DF++ is built upon our earlier Celeb-DF dataset~\cite{li2020celeb}, containing {$59$} celebrities of varying genders, ages, and ethnicities. The real video set remains unchanged, which includes $590$ videos publicly available on YouTube. Using numerous advanced and latest DeepFake methods, we expand the DeepFake videos to $53196$, each lasting on average $10$ seconds and totaling over $15$ million frames. Following the widely used evaluation setup in previous works~\cite{zhao2021multi,yan2023ucf,cui2024forensics}, we demonstrate the increased difficulty of the proposed benchmark. The details of this benchmark are shown in Table~\ref{table:datasets}.

To comprehensively measure the generalizability of DeepFake detection methods, we establish three challenging evaluation protocols relying on the Celeb-DF++ benchmark: \textit{Generalized Forgery} evaluation (GF-eval), \textit{Generalized Forgery across Quality} evaluation (GFQ-eval), and \textit{Generalized Forgery across Datasets} evaluation (GFD-eval). GF-eval assesses detection performance across DeepFake methods in all three forgery scenarios, reflecting real-world conditions where forgeries are diverse and potentially unseen. GFQ-eval increases the challenge by evaluating the detection performance under varying video compression levels. This aligns with the reality that online videos are often subject to compression, and this compression can obscure forgery traces, raising the challenge for generalizable forensics. GFD-eval measures the generalizability of detection methods across different datasets, simulating the practical case where training and testing data originate from different sources.

With these evaluation protocols, we conduct extensive experiments using \textbf{24} recent detection methods. Compared to existing datasets, our evaluation incorporates more recent methods, highlighting its comprehensiveness and timeliness.
Results reveal that generalizable DeepFake detection remains an open challenge, calling for sustained research and innovation.


\begin{table*}[!ht]
    \centering
    \caption{\small Comparison of DeepFake video benchmarks.
    We highlight that Celeb-DF++ is more diverse, including \textbf{22} various DeepFake methods that span across Face-Swap (FS), Face-Reenactment (FR), and Talking Face (TF) scenarios, and conduct comprehensive up-to-date evaluations using \textbf{24} detectors, including \textbf{5} recent ones released after 2024.}
    \vspace{-0.3cm}
    \label{table:datasets}
    \setlength{\tabcolsep}{3pt}
    \resizebox{1.0\linewidth}{!}{
    \begin{tabular}{l|l|cc|cccc|c|ccc|c|c|c|cc|c}
    \hline
    \multirow{2}{*}{Dataset}
    & \multirow{2}{*}{Venue}
    & \multicolumn{2}{c|}{Modality} 
    & \multicolumn{4}{c|}{DeepFake Model}
    & \multicolumn{4}{c|}{DeepFake Method}
    & \multirow{2}{*}{\#Real}
    & \multirow{2}{*}{\#Fake}
    & \multicolumn{3}{c|}{Assessed Detectors}
    & \multirow{2}{*}{Source}\\
        \cline{3-12}
        \cline{15-17}
    && Visual & Audio & AE&GAN & DM & Other\tablefootnote{Other category includes traditional graphics-based approaches~\cite{nirkin2018face, faceswap}, neural rendering technique~\cite{thies2019deferred}, and {3DMM–based generation methods}~\cite{wang2021hififace, thies2016face2face}.}  & \#Total
    & \#FS
    & \#FR
    & \#TF & & & \#Total & Pre'24 & Post'24  & \\

    \hline
    DF-TIMIT\cite{korshunov2018deepfakes} & ArXiv'18 & \ding{51} & && \ding{51} & & & 2 & 2 & - & - & 320 & 640 & 4 & 4 & - & \href{https://www.idiap.ch/en/scientific-research/data/deepfaketimit}{Link}\\
    DFDCP\cite{dolhansky2019dee} & ArXiv'19 & \ding{51} & && \ding{51} & & & 2 & 2 & - & - & 1,131 & 4,113 & 3 & 3 & - & \href{https://ai.meta.com/datasets/dfdc/}{Link}\\
    UADFV\cite{yang2018exposing} & ICASSP'19 & \ding{51} && & & & \ding{51} & 1 & 1 & - & - & 49 & 49 & 6 & 6 & - & \href{https://github.com/yuezunli/WIFS2018_In_Ictu_Oculi}{Link}\\
    FF++\cite{rossler2019faceforensics++} & ICCV'19 & \ding{51} & & \ding{51}& & & \ding{51} & 4 & 2 & 2 & - & 1,000 & 4,000 & 6 & 6 & - & \href{https://github.com/ondyari/FaceForensics}{Link}\\
    DFD\cite{DDD_GoogleJigSaw2019} & - & \ding{51} & & -& - & - & - & 5 & 5 & - & - & 363 & 3,068 & - & - & - & \href{https://research.google/blog/contributing-data-to-deepfake-detection-research/}{Link}\\
    DFDC\cite{dolhansky2020deepfake} & ArXiv'20 & \ding{51} & & \ding{51} &\ding{51}& & \ding{51} & 8 & 6 & - & 2 & 23,654 & 104,500 & 5 & 5 & - & \href{https://ai.meta.com/datasets/dfdc/}{Link}\\
    DFFD\cite{dang2020detection} & CVPR'20 & \ding{51} & & \ding{51} &\ding{51}& & \ding{51}& 7 & 7 & - & - & 1,000 & 3,000 & 10 & 10 & - & \href{http://cvlab.cse.msu.edu/project-ffd.html}{Link}\\
    DForensics-1.0\cite{jiang2020deeperforensics} & CVPR'20 & \ding{51} & & \ding{51} && & & 1 & 1 & - & - & 50,000 & 10,000 & 5 & 5 & - & \href{https://github.com/EndlessSora/DeeperForensics-1.0}{Link}\\
    WildDeepfake\cite{zi2020wilddeepfake} & MM'20 & \ding{51} & & - & - & - & - &-& - & - & - & 3,805 & 3,509 & 15 & 15 & - & \href{https://github.com/OpenTAI/wild-deepfake}{Link}\\

    OpenForensics\tablefootnote{Note that OpenForensics~\cite{le2021openforensics} only contains images, \#Real and \#Fake in the table refer to the number of images rather than videos.}\cite{le2021openforensics} & ICCV'21 & \ding{51} & && \ding{51} & & & 1 & 1 & - & - & 45,473 & 70,325 & 12 & 12 & - & \href{https://sites.google.com/view/ltnghia/research/openforensics/}{Link}\\
    KoDF\cite{kwon2021kodf} & ICCV'21 & \ding{51} & & \ding{51} & \ding{51}& &\ding{51} & 6 & 3 & 1 & 2 & 62,166 & 175,776 & 1 & 1 & - &\href{https://deepbrainai-research.github.io/kodf/}{Link}\\
    FFIW\cite{zhou2021face} & CVPR'21 & \ding{51} & & \ding{51} &  \ding{51} &&& 3 & 3 & - & - & 10,000 & 10,000 & 9 & 9 & - & \href{https://github.com/tfzhou/FFIW}{Link}\\
    FakeAVCeleb\cite{khalid2021fakeavceleb} & NeurIPS'21 & \ding{51} & \ding{51} &\ding{51}& \ding{51} & & \ding{51} & 4 & 2 & - & 2 & 500 & 19,500 & 8 & 8 & - & \href{https://github.com/DASH-Lab/FakeAVCeleb}{Link}\\

    DFDM\cite{jia2022model} & ICIP'22 & \ding{51} & &\ding{51}  && &  & 5 & 5 & - & - & 590 & 6,450 & 9 & 9 & - & \href{https://github.com/shanface33/Deepfake_Model_Attribution}{Link}\\
    
    DF-Platter\cite{narayan2023df} & CVPR'23 & \ding{51} & & \ding{51}&\ding{51} & &  & 3 & 3 & - & - & 764 & 132,496 & 6 & 6 & - & \href{https://iab-rubric.org/df-platter-database}{Link}\\
    DefakeAVMiT\cite{yang2023avoid} & TIFS'23 & \ding{51} & \ding{51} & \ding{51} & \ding{51}& \ding{51} && 5 & 2 & - & 3 & 540 & 6,480 & 21 & 21 & - & \href{https://github.com/SYSU-DISG/AVoiD-DF}{Link}\\
    AV-Deepfake1M\cite{cai2024av} & MM'24 & \ding{51} & \ding{51} & & \ding{51} &&& 1 & - & - & 1 & 286,721 & 860,039 & 22 & 22 & - & \href{https://github.com/ControlNet/AV-Deepfake1M}{Link}\\
    THBench\cite{xiong2025talkingheadbench} & ArXiv'25 & \ding{51} & \ding{51} & & \ding{51} & \ding{51} && 6 & - & - & 6 & 2,312 & 2,984 & 4 & 2 & 2 & \href{https://huggingface.co/datasets/luchaoqi/TalkingHeadBench}{Link} \\
    \hline
    (Ours) Celeb-DF\cite{li2020celeb} & CVPR'20 & \ding{51} & & \ding{51} & & && 1 & 1 & - & - & 590 & 5,639 & 13 & 13 & - & \href{https://github.com/yuezunli/celeb-deepfakeforensics}{Link} \\
    \rowcolor{hl} (Ours) \textbf{Celeb-DF++} & - & \ding{51} & \ding{51} & \ding{51} & \ding{51} & \ding{51}& \ding{51} & \textbf{22} & \textbf{8} & \textbf{7} & \textbf{7} & 590 & 53,196 & \textbf{24} & 19 & \textbf{5} & \href{https://github.com/OUC-VAS/Celeb-DF-PP}{Link} \\
    \hline
    
    \end{tabular}
    }

\end{table*}

\section{Related Works}
\subsection{DeepFake Generation}

\smallskip
\noindent\textbf{Original DeepFake}.
The term DeepFake emerged in 2017, initially referring to face-swap forgery techniques. The typical pipeline involves extracting faces from a source video and feeding them into autoencoder-based generative models, which synthesize the faces of a target individual with the same facial attributes, such as orientation and expressions. Over the past few years, face-swap techniques have been well-developed, encouraging the release of many open-source tools, \eg,~\cite{fakeapp,DFaker,faceswap-gan,faceswap,DeepFaceLab,facefusion,3dfaceswap}. While their methodology is relatively simple, they are user-friendly, perform reliably, and have gained widespread popularity. To date, a significant proportion of DeepFake videos circulating online are produced using these tools, shaping the current landscape of face-swap forgeries.   

\smallskip
\noindent\textbf{Modern DeepFake}. With the rapid evolution of generative models, face-swap techniques become increasingly efficient and effective, constantly improving the generated qualities,~\eg,~\cite{chen2020simswap,shiohara2023blendface}. Alongside this progress, many other categories of media forgery have emerged. Among them, Face-reenactment and Talking-face have recently attracted considerable attention~\cite{rochow2024fsrt,guo2024liveportrait,zhang2023sadtalker,liu2024anitalker}. Different from face-swap, face-reenactment generates entire video frames based on the driving source, while Talking-face generates the lip movement and even facial emotions conditioned on given audio tracks. Other forgery categories, such as facial attribute editing~\cite{jiang2025towards,huang2024sdgan} and style transfer~\cite{zhou2024deformable,xu2025stylessp}, are comparatively less deceptive, as they only focus on manipulating images and struggle to maintain consistency across video frames. Interestingly, generative models have also enabled audio-based forgeries, such as synthesizing speech in the voice of a target individual who never actually speak these words~\cite{du2024cosyvoice,chen2024f5}. 
As a result, the concept of DeepFake has broadened to stand for a wide range of AI-based forgery scenarios. 


\subsection{DeepFake Detection}
\smallskip
\noindent\textbf{Conventional Detection}. 
To identify DeepFakes, a large amount of detection methods have been proposed in a short period~\cite{li2018ictu,yang2018exposing,qi2020deeprhythm,li2019exposing,matern2019exploiting,sun2022information, qian2020thinking,dang2020detection}. Most of these methods are based on deep neural networks (DNNs), leveraging their strong feature learning capacities. Based on the types of features they aim to extract, these methods can be roughly divided into several categories. One classical category focuses on detecting inconsistencies in physical or physiological characteristics, such as abnormal eye blinking~\cite{li2018ictu}, irregular head pose movement~\cite{yang2018exposing}, or disrupted heart beat rhythms~\cite{qi2020deeprhythm}. Another common direction attempts to capture forgery artifacts introduced by the generation process, including blending artifacts~\cite{li2019exposing,matern2019exploiting}, generative artifacts~\cite{qian2020thinking,luo2023beyond}, and temporal artifacts~\cite{guo2025face,yu2025mining}, in either spatial or frequency domains. A third line of work designs specialized network architectures and training objectives to directly learn discriminative features from data. These include strategies incorporating attention mechanism~\cite{sun2022information}, contrastive learning~\cite{sun2022dual}, enhanced feature extraction module~\cite{dang2020detection}, etc.

\smallskip
\noindent\textbf{Generalizable Detection}. 
Despite showing promising results, detection methods struggle to detect unseen forgeries, limiting their practicalness. Many effort has been devoted to address this by enhancing detection generalizability. One effective solution is to create diverse pseudo-fake training faces, by simulating the forgery artifacts commonly observed in real DeepFakes~\cite{shiohara2022detecting,zhou2024freqblender,cheng2024can,cui2024forensics}. Another line of research focuses on learning generalizable forgery features from known training samples, using strategies such as adversarial training~\cite{chen2022self}, feature disentanglement~\cite{yan2023ucf}, self-supervised learning~\cite{li2023spatio}. While these methods have improved the generalizability, the lack of diversity in existing datasets hinder further advancements in this field.

\subsection{DeepFake Datasets}
\label{sec:prev-dbs}
The training and evaluation of DeepFake detection methods require datasets. Early DeepFake datasets were generally limited in scale due to the incapacity of initial DeepFake methods. Some of the first publicly available datasets, such as {UADFV}~\cite{li2018ictu} and {DeepFake-TIMIT (DF-TIMIT)}~\cite{korshunov2018deepfakes}, contains only a small number of DeepFake videos generated using earlier face-swap methods~\cite{fakeapp,faceswap-gan}. 
The release of {FaceForensics++ (FF++)}~\cite{rossler2019faceforensics++} marks the emergence of large-scale DeepFake datasets, including thousands of real videos and the same number of DeepFake videos generated using various face-swap manipulations. Later, Google \& JigSaw releases {DeepFakeDetection (DFD)} dataset~\cite{DDD_GoogleJigSaw2019}, featuring a larger collection of DeepFake videos, sourced from consented actors. 

With recent advancement in DeepFakes, datasets have continued to evolve, significantly increasing in scale and diversity. Facebook (now Meta) launches a {DeepFake Detection Challenge (DFDC)} and releases a corresponding dataset~\cite{dolhansky2020deepfake}. Initially, a preview version (DFDCP) is released and upon completion of the challenge, the full dataset is made publicly available. This dataset contained hundreds of thousands of video clips featuring paid actors and used various GAN-based and non-learning-based DeepFake methods. Other mainstream datasets such as {DFFD}~\cite{dang2020detection}, {DeeperForensics-1.0 (DForensics-1.0)}~\cite{jiang2020deeperforensics}, and {WildDeepfake}~\cite{zi2020wilddeepfake}, are timely released with large scale, covering more diverse scenes and a wider range of facial expressions. Our prior work, Celeb-DF dataset~\cite{li2020celeb}, includes larges-scale and challenging DeepFake videos of celebrities, featuring enhanced visual quality achieved by a curated generation pipeline. Moreover, a number of specialized datasets have been created to address specific challenges, such as the lack of Asian subjects in {KoDF} dataset~\cite{kwon2021kodf}, the model attribution problem in  {DFDM} dataset~\cite{jia2022model}, the detection of talking heads in {TalkingHeadBench (THBench)} dataset~\cite{xiong2025talkingheadbench}, and multi-face forgery detection in {OpenForensics} dataset~\cite{le2021openforensics}, {FFIW} dataset~\cite{zhou2021face} and {DF-Platter} dataset~\cite{narayan2023df}.

In addition to unimodal datasets, several multimodal datasets have emerged, combining both visual and audio information, such as {FakeAVCeleb} dataset~\cite{khalid2021fakeavceleb}, {DefakeAVMiT}~\cite{yang2023avoid} and {AV-Deepfake1M} dataset~\cite{cai2024av}.

We highlight that while existing datasets offer promising data scales, they lack sufficient diversity in DeepFake methods. Most of them only consider a single forgery scenario such as face-swap, limiting their ability to effectively assess the generalizability of detection methods. Detailed comparison is shown in Table~\ref{table:datasets}.

\section{The Celeb-DF++ Benchmark}
Celeb-DF++ is extended from our previous Celeb-DF dataset~\cite{li2020celeb} with more diversity, including \textbf{22} various DeepFake methods that span across Face-swap (FS), Face-reenactment (FR), and Talking-face (TF) scenarios. Moreover, we conduct comprehensive up-to-date evaluations using \textbf{24} detectors, including \textbf{5} recent ones released after 2024 that have not been considered in existing datasets. A summary of existing datasets are shown in Table~\ref{table:datasets}.

\begin{table}[!t]
    \centering
    \caption{\small {DeepFake methods used in Celeb-DF++ benchmark.}}
    \vspace{-0.3cm}
    \label{tab:involved_deepfake}
    \begin{tabular}{c|l|l|l}
    \hline
    Scenario & DeepFake Method & Venue & Code \\
    \hline
        \multirow{8}{*}{FS}
        & Celeb-DF\cite{li2020celeb} & CVPR'20 & \href{https://github.com/yuezunli/celeb-deepfakeforensics}{Link} \\
        & SimSwap\cite{chen2020simswap} & MM'20 & \href{https://github.com/neuralchen/SimSwap}{Link} \\
        & InSwapper\cite{inswapper} & - & \href{https://github.com/haofanwang/inswapper}{Link} \\
        & HifiFace\cite{wang2021hififace} & IJCAI'21 & \href{https://github.com/maum-ai/hififace}{Link} \\        
        & GHOST\cite{groshev2022ghost} & Access'22 & \href{https://github.com/ai-forever/ghost}{Link} \\                
        & UniFace\cite{xu2022designing} & ECCV'22 & \href{https://github.com/xc-csc101/UniFace}{Link} \\                
        & MobileFaceSwap\cite{xu2022mobilefaceswap} & AAAI'22 & \href{https://github.com/Seanseattle/MobileFaceSwap}{Link} \\                
        & BlendFace\cite{shiohara2023blendface} & ICCV'23 & \href{https://github.com/mapooon/BlendFace}{Link} \\  
    \hline
        \multirow{7}{*}{FR}
        & DaGAN\cite{hong2022depth} & CVPR'22 & \href{https://github.com/harlanhong/CVPR2022-DaGAN}{Link} \\
        & TPSMM\cite{zhao2022thin} & CVPR'22 & \href{https://github.com/yoyo-nb/Thin-Plate-Spline-Motion-Model}{Link} \\
        & MCNET\cite{hong2023implicit} & ICCV'23 & \href{https://github.com/harlanhong/ICCV2023-MCNET}{Link} \\
        & HyperReenact\cite{bounareli2023hyperreenact} & ICCV'23 & \href{https://github.com/StelaBou/HyperReenact}{Link} \\        
        & LIA\cite{wang2024lia} & TPAMI'24 & \href{https://github.com/wyhsirius/LIA}{Link} \\                
        & FSRT\cite{rochow2024fsrt} & CVPR'24 & \href{https://github.com/andrerochow/fsrt}{Link} \\                
        & LivePortrait\cite{guo2024liveportrait} & ArXiv'24 & \href{https://github.com/KwaiVGI/LivePortrait}{Link} \\            
    \hline
        \multirow{7}{*}{TF}
        & SadTalker\cite{zhang2023sadtalker} & CVPR'23 & \href{https://github.com/OpenTalker/SadTalker}{Link} \\
        & IP-LAP\cite{zhong2023identity} & CVPR'23 & \href{https://github.com/Weizhi-Zhong/IP_LAP}{Link} \\
        & AniTalker\cite{liu2024anitalker} & MM'24 & \href{https://github.com/X-LANCE/AniTalker}{Link} \\        
        & EDTalk\cite{tan2024edtalk} & ECCV'24 & \href{https://github.com/tanshuai0219/EDTalk}{Link} \\             
        & Real3D-Portrait\cite{ye2024real3d} & ICLR'24 & \href{https://github.com/yerfor/Real3DPortrait}{Link} \\  

        & EchoMimic\cite{chen2025echomimic} & AAAI'25 & \href{https://github.com/antgroup/echomimic}{Link} \\    
        & FLOAT\cite{ki2024float} & ICCV'25 & \href{https://github.com/deepbrainai-research/float}{Link} \\  
    \hline
    \end{tabular}
\end{table}

\subsection{Revisit of Celeb-DF Dataset}
The Celeb-DF dataset comprises {$590$} real videos and {$5,639$} DeepFake videos (corresponding to over two million video frames). The average length of all videos is approximate $13$ seconds with the standard frame rate of $30$ frame-per-second. The real videos are chosen from publicly available {YouTube} videos, corresponding to interviews of {$59$} celebrities with a diverse distribution in their genders, ages, and ethnic groups\footnote{We choose celebrities' faces as they are more familiar to the viewers so that any visual artifacts can be more readily identified. Furthermore, celebrities are anecdotally the main targets of DeepFake videos.}. $56.8\%$ subjects in the real videos are male, and $43.2\%$ are female. $8.5\%$ are of age 60 and above, $30.5\%$ are between 50 - 60, $26.6\%$ are 40s, $28.0\%$ are 30s, and $6.4\%$ are younger than 30. $5.1\%$ are Asians, $6.8\%$ are African Americans and $88.1\%$ are Caucasians. In addition, the real videos exhibit large range of changes in aspects such as the subjects' face sizes (in pixels), orientations, lighting conditions, and backgrounds. The DeepFake videos are generated by swapping faces for each pair of the {$59$} subjects. The final videos are in MPEG4.0 format.

To generate DeepFake videos, we improve the initial face-swap method~\cite{fakeapp} by enlarging the synthesis size, reducing color mismatch between swapped area and surroundings, refining the swapping masks, and mitigating temporal flickering. This dataset has been commonly used for training and evaluating DeepFake detection models. However, due to limited diversity of released DeepFake methods, this dataset only involves face-swap scenario with a single DeepFake method. Thus, it is insufficient to assess the generalizability of DeepFake detection methods.

\subsection{DeepFake Methods and Generation Details}
\begin{figure*}[t]
\centering
\includegraphics[width=\linewidth]{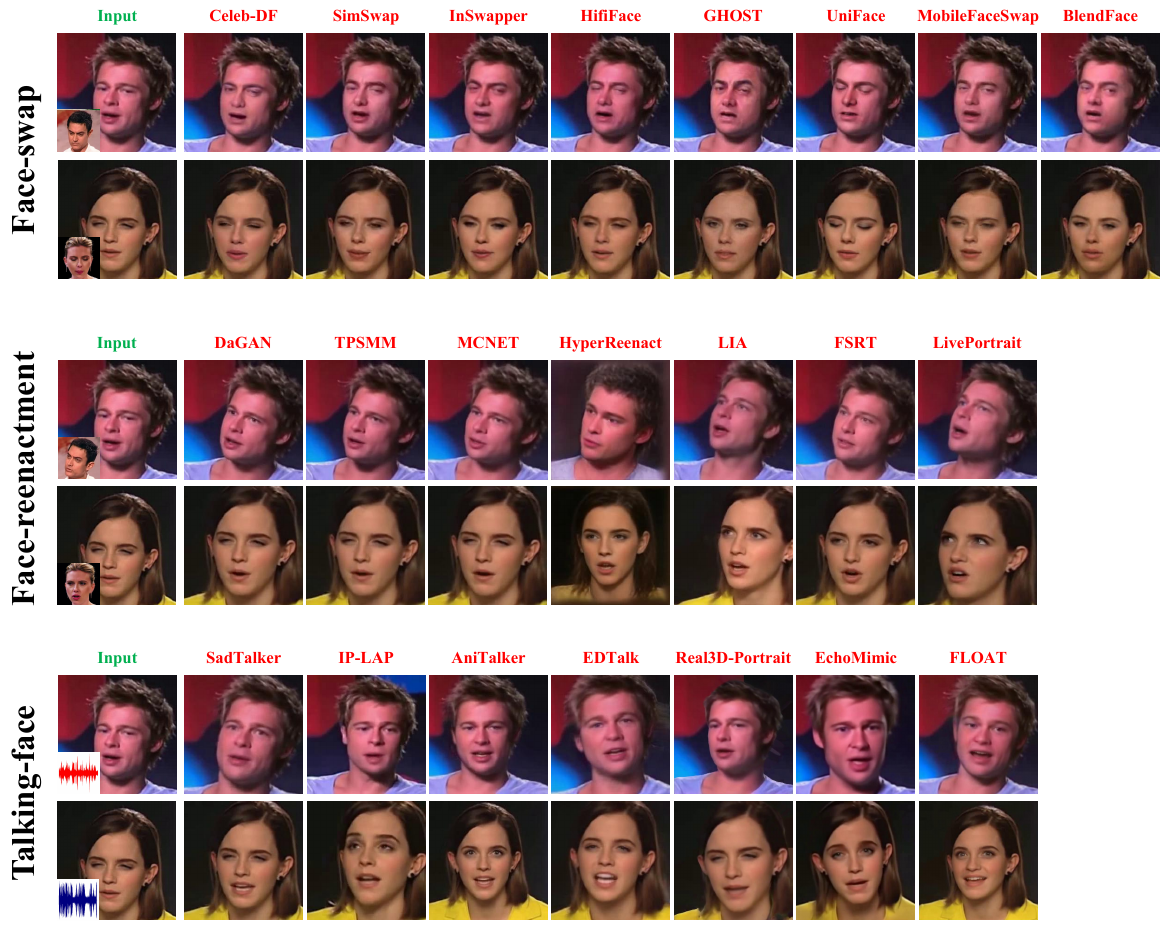}
\vspace{-0.7cm}
\caption{\small {Visual examples of Celeb-DF++. The first column denotes the source identity face and the bottom-left embedded image represents the target face or the driving face/audio. The remaining columns display the results generated by each DeepFake method.}}
\label{fig:vis2}
\end{figure*}

In Celeb-DF++, we explore three forgery scenarios: Face-swap (FS), Face-reenactment (FR) and Talking-face (TF), respectively, and each scenario is built using a large number of recent DeepFake methods. Table~\ref{tab:involved_deepfake} illustrates the details of each method, and Fig.~\ref{fig:vis2} shows several generation examples.

\begin{table}[!t]
    \centering
    \caption{\small Assessed DeepFake Detection Methods.}
    \vspace{-0.3cm}
    \label{tab:detection-methods}
    \resizebox{\linewidth}{!}{
    \begin{tabular}{l|l|l|l|l}
    \hline
    Detector & Venue & Architecture & Category\tablefootnote{Since detection methods often rely on a combination of multiple clues, we report only the most representative clue to define their category.} & Code \\
    \hline
        MesoNet\cite{afchar2018mesonet} & WIFS'18 & Designed CNN & Data-driven & \href{https://github.com/HongguLiu/MesoNet-Pytorch}{Link} \\     
        MesoInception\cite{afchar2018mesonet} & WIFS'18 & Designed CNN & Data-driven & \href{https://github.com/HongguLiu/MesoNet-Pytorch}{Link} \\
        Xception\cite{rossler2019faceforensics++} & ICCV'19 & Xception & Data-driven & \href{https://github.com/ondyari/FaceForensics}{Link} \\
        EfficientNet-B4\cite{tan2019efficientnet} & ICML'19 & EfficientNet & Data-driven & \href{https://github.com/tensorflow/tpu/tree/master/models/official/efficientnet}{Link} \\ 
        Capsule\cite{nguyen2019capsule} & ICASSP'19 & CapsuleNet & Data-driven & \href{https://github.com/nii-yamagishilab/Capsule-Forensics-v2}{Link} \\
        F3Net\cite{qian2020thinking} & ECCV'20 & Designed CNN & Frequency & \href{https://github.com/yyk-wew/F3Net}{Link} \\
        CNN-Aug\cite{wang2020cnn} & CVPR'20 & ResNet  & Data-driven & \href{https://github.com/peterwang512/CNNDetection}{Link} \\
        FFD\cite{dang2020detection} & CVPR'20 & Designed CNN  & Attention-driven & \href{https://github.com/JStehouwer/FFD_CVPR2020}{Link} \\
        SPSL\cite{liu2021spatial} & CVPR'21 & Designed CNN  & Frequency & \href{https://github.com/SCLBD/DeepfakeBench}{Link} \\
        SRM\cite{luo2021generalizing} & CVPR'21 & Designed CNN  & Frequency & \href{https://github.com/crywang/face-forgery-detection}{Link} \\
        RFM\cite{wang2021representative} & CVPR'21 & Designed CNN  & Attention-driven & \href{https://github.com/crywang/RFM}{Link} \\
        MATT\cite{zhao2021multi} & CVPR'21& Designed CNN  & Attention-driven  & \href{https://github.com/yoctta/multiple-attention}{Link} \\
        CLIP\cite{radford2021learning} & ICML'21 & CLIP & Data-driven  & \href{https://github.com/OpenAI/CLIP}{Link} \\
        RECCE\cite{cao2022end} & CVPR'22 & Designed CNN  & Reconstruction & \href{https://github.com/VISION-SJTU/RECCE}{Link} \\
        SBI\cite{shiohara2022detecting} &CVPR'22 &Designed CNN & Pseudo-fake & \href{https://github.com/mapooon/SelfBlendedImages}{Link}\\
        CORE\cite{ni2022core} & CVPRW'22 & Designed CNN  & Feature consistency & \href{https://github.com/niyunsheng/CORE}{Link} \\ 
        SIA\cite{sun2022information} & ECCV'22 & Designed CNN  & Attention-driven  & \href{https://github.com/skJack/Code-for-SIA}{Link} \\
        UCF\cite{yan2023ucf} & ICCV'23 & Designed CNN  & Disentangle & \href{https://github.com/SCLBD/DeepfakeBench}{Link} \\
        IID\cite{huang2023implicit} & CVPR'23 & Designed CNN  & Identity discrepancy & \href{https://github.com/SCLBD/DeepfakeBench}{Link} \\
        LSDA\cite{yan2024transcending} & CVPR'24 & Designed CNN  & Augmentation & \href{https://github.com/SCLBD/DeepfakeBench}{Link} \\
        CFM\cite{luo2023beyond} & TIFS'24 & Designed CNN & Fine-grained & \href{https://github.com/LoveSiameseCat/CFM}{Link} \\
        ProDet\cite{cheng2024can} & NeurIPS'24 & Designed CNN  & Pseudo-fake & \href{https://github.com/beautyremain/ProDet}{Link} \\
        ForAda\cite{cui2024forensics} & CVPR'25 & Designed CLIP & Adaptation & \href{https://github.com/OUC-VAS/ForensicsAdapter}{Link} \\
        Effort\cite{yan2024effort} & ICML'25 & Designed CLIP & Adaptation & \href{https://github.com/YZY-stack/Effort-AIGI-Detection}{Link} \\
    \hline 
    \end{tabular}
    }
\end{table}
\smallskip
\noindent\textbf{Face-swap Scenario.}
This scenario denotes the face forgery where the original face area (\eg, whole face or facial organs such as eye, mouth, etc) is replaced by a newly synthesized face area, while retaining the same behavioral attributes. We include 8 methods in this scenario, which are the improved face-swap method in Celeb-DF~\cite{li2020celeb}, and seven additional methods: SimSwap~\cite{chen2020simswap}, InSwapper~\cite{inswapper}, HifiFace~\cite{wang2021hififace}, GHOST~\cite{groshev2022ghost}, UniFace~\cite{xu2022designing}, MobileFaceSwap~\cite{xu2022mobilefaceswap}, BlendFace~\cite{shiohara2023blendface}. These methods rely on auto-encoder or GAN, with different learning strategies such as ID-preserving module (SimSwap), roop-based tool (InSwapper), 3D shape-aware identity extractor (HifiFace), blending/stabilization mechanism (GHOST), identity-consistent reconstruction (UniFace), knowledge distillation (MobileFaceSwap), and disentanglement-based identity guidance (BlendFace). The data scale for each additional method is greater than $1900$. 

\smallskip
\noindent\textbf{Face-reenactment Scenario.}
Under this, entirely new videos are generated featuring a target individual whose facial expressions, movements, and behaviors are driven by the behaviors of a source individual. This scenario includes 7 recent methods, including DaGAN~\cite{hong2022depth}, TPSMM~\cite{zhao2022thin}, MCNET~\cite{hong2023implicit}, HyperReenact~\cite{bounareli2023hyperreenact}, LIA~\cite{wang2024lia}, FSRT~\cite{rochow2024fsrt}, LivePortrait~\cite{guo2024liveportrait}. These methods utilize various strategies, such as depth-based 3D face reconstruction (DaGAN), motion-estimation–driven optical flow generation (TPSMM), implicit identity memory network (MCNET), hypernetwork-based image inversion (HyperReenact), latent-space linear motion modeling (LIA), Transformer-based latent representation learning (FSRT), and keypoint-guided controllability (LivePortrait). Each method generates more than $1900$ DeepFake videos.

\smallskip
\noindent\textbf{Talking-face Scenario.}
This scenario involves using audio input to create face video of a target individual, with the lip movements and emotions flawlessly aligning with the audio. To build this scenario, we adopt 7 methods, including SadTalker~\cite{zhang2023sadtalker}, IP-LAP~\cite{zhong2023identity}, AniTalker~\cite{liu2024anitalker}, EDTalk~\cite{tan2024edtalk}, Real3D-Portrait~\cite{ye2024real3d}, EchoMimic~\cite{chen2025echomimic}, FLOAT~\cite{ki2024float}. These methods are diverse in generation strategies, such as using audio-driven 3DMM expression modeling (SadTalker), landmark- and appearance-guided synthesis (IP\_LAP), generic motion representation learning (AniTalker), component-disentangled training (EDTalk), one-shot 3D reconstruction via image-to-plane modeling (Real3D-Portrait), joint audio-landmark supervision (EchoMimic), and flow-based generation via motion field matching (LivePortrait). Each method results in more than $2700$ DeepFake videos.

\smallskip
\noindent\textbf{Data Generation Details.}
For Face-swap and Face-reenactment scenario, we randomly select $2,000$ identity pairs from real video set to generate DeepFake videos,  resulting in $13,646$ videos, excluding those from the previous Celeb-DF dataset. For the Talking-face scenario, we use the first frame of each real video as the source and randomly select $5$ audio segments from the VoxCeleb2 dataset~\cite{chung2018voxceleb2} for each video frame to drive the synthesis, generating $20,279$ DeepFake videos. In total, $53,196$ DeepFake videos are generated.

\smallskip
\noindent\textbf{Training and Testing Split.}
For real videos, we follow the split of Celeb-DF, where $178$ videos are selected. For DeepFake videos, we randomly select $200$ videos per method in the Face-swap scenario, $200$ videos per method in the Face-reenactment scenario, and $300$ videos per method in the Talking-face scenario, respectively. More details can be found on the project page.

\subsection{Assessed DeepFake Detection Methods}
We incorporate 24 mainstream DeepFake detectors for assessment, including MesoNet\cite{afchar2018mesonet} (and its variant MesoInception), Xception\cite{rossler2019faceforensics++}, EfficientNet-B4\cite{tan2019efficientnet}, Capsule\cite{nguyen2019capsule}, F3Net\cite{qian2020thinking}, CNN-Aug\cite{wang2020cnn}, FFD\cite{dang2020detection}, SPSL\cite{liu2021spatial}, SRM\cite{luo2021generalizing}, RFM\cite{wang2021representative}, CLIP\cite{radford2021learning}, Multi-attention (MATT)\cite{zhao2021multi}, RECCE\cite{cao2022end}, SBI\cite{shiohara2022detecting}, CORE\cite{ni2022core}, SIA\cite{sun2022information}, UCF\cite{yan2023ucf}, IID\cite{huang2023implicit}, LSDA\cite{yan2024transcending}, CFM\cite{luo2023beyond}, ProDet\cite{cheng2024can}, ForensicsAdapter (ForAda)\cite{cui2024forensics}, and Effort\cite{yan2024effort}. All deepfake detectors are trained and tested using their default settings. Notably, most of the methods are implemented following the default settings and pre-processing procedures in DeepfakeBench~\cite{yan2023deepfakebench}. For the most recent methods, such as CFM, ProDet, ForAda, and Effort, we rigorously implement them according to their official code. A summarization is shown in Table~\ref{tab:detection-methods}. 

Table~\ref{table:datasets} shows the comparison of evaluation scale among different datasets. From the statistics, we highlight that our benchmark involves more recent detection methods, better reflecting the current advancements in the field and uncovering the potential direction in the future.

To assess these detection methods, we employ frame-level and video-level ROC AUC metrics, which are commonly adopted in detection tasks, \eg, \cite{qian2020thinking,zhao2021multi,cui2024forensics}. Specifically, we randomly sample $32$ frames for each tested video. For the frame-level AUC, we obtain the detection probability for each sampled frame, and calculate the AUC score using all sampled frames. For the video-level AUC, we first average the probabilities of the sampled frames within each video to obtain a video-level probability, and calculate AUC score using these probabilities.

\begin{table}[!t]
    \centering
    \caption{\small \textbf{Frame-level AUC} ($\%$) evaluation results. All models are trained on FF++ (HQ) and tested on other datasets.}
    \vspace{-0.3cm}
    \label{tab:frame3}
    \begin{tabular}{l|l|c|c|>{\columncolor{hl}}c}
    \hline
    \multirow{2}{*}{Detector} & 
    \multirow{2}{*}{Venue} & \multicolumn{2}{c|}{Celeb-DF~\cite{li2020celeb}} & \\
    \cline{3-4}
    & & v1 & v2 & \multirow{-2}{*}{\textbf{Celeb-DF++}} \\
    \hline
        MesoNet\cite{afchar2018mesonet} & WIFS'18  & 49.5 & 53.1 & 48.3 \\     
        MesoInception\cite{afchar2018mesonet} & WIFS'18 & 69.3 & 65.0 & 64.0 \\
        Xception\cite{rossler2019faceforensics++} & ICCV'19 & 75.4 & 74.0 & 73.7 \\
        EfficientNet-B4\cite{tan2019efficientnet} & ICML'19 & 76.7 & 75.0 & 70.6 \\ 
        Capsule\cite{nguyen2019capsule} & ICASSP'19 & 77.3 & 76.5 & 70.9 \\
        F3Net\cite{qian2020thinking} & ECCV'20 & 75.0 & 72.9 & 70.2 \\
        CNN-Aug\cite{wang2020cnn} & CVPR'20 & 71.5 & 67.5 & 61.7 \\
        FFD\cite{dang2020detection} & CVPR'20 & 70.9 & 68.7 & 67.1 \\
        SPSL\cite{liu2021spatial} & CVPR'21 & 80.4 & 72.9 & 68.4 \\
        SRM\cite{luo2021generalizing} & CVPR'21 & 76.5 & 76.0 & 72.6 \\
        RFM\cite{wang2021representative} & CVPR'21 & 79.3 & 76.4 & 70.4 \\
        MATT\cite{zhao2021multi} & CVPR'21 & 75.5 & 72.0 & 67.0 \\
        CLIP\cite{radford2021learning} & ICML'21 & 85.7 & 82.7 & 75.1 \\
        RECCE\cite{cao2022end} & CVPR'22 & 73.4 & 74.1 & 75.5 \\
        SBI\cite{shiohara2022detecting} & CVPR'22 & 69.1 & 74.8 & 70.9 \\
        CORE\cite{ni2022core} & CVPRW'22 & 71.8 & 74.1 & 70.4 \\ 
        SIA\cite{sun2022information} & ECCV'22 & 81.5 & 72.6 & 65.0 \\
        UCF\cite{yan2023ucf} & ICCV'23 & 81.1 & 77.2 & 72.4 \\
        IID\cite{huang2023implicit} & CVPR'23 & 73.0 & 74.7 & 71.4 \\
        LSDA\cite{yan2024transcending} & CVPR'24 & 74.7 & 73.7 & 70.0 \\
        CFM\cite{luo2023beyond} & TIFS'24 & 83.6 & 81.1 & 73.3 \\
        ProDet\cite{cheng2024can} & NeurIPS'24 & 87.6 & 84.2 & 69.2 \\
        ForAda\cite{cui2024forensics} & CVPR'25 & 91.4 & 89.9 & 71.7 \\
        Effort\cite{yan2024effort} & ICML'25 & 86.4 & 86.8 & 80.8 \\
    \hline 
    Average & - & 76.5 & 74.8 & 69.6 \\
    \hline
    \end{tabular}
\end{table}
\begin{table}[!t]
    \centering
    \caption{\small \textbf{Video-level AUC} ($\%$) evaluation results. All models are trained on FF++ (HQ) and tested on other datasets.}
    \vspace{-0.3cm}
    \label{tab:video3}
    \begin{tabular}{l|l|c|c|>{\columncolor{hl}}c}
    \hline
    \multirow{2}{*}{Detector} & 
    \multirow{2}{*}{Venue} & \multicolumn{2}{c|}{Celeb-DF~\cite{li2020celeb}} & \\
    \cline{3-4}
    & & v1 & v2 & \multirow{-2}{*}{\textbf{Celeb-DF++}} \\
    \hline 
        MesoNet\cite{afchar2018mesonet} & WIFS'18 & 49.4 & 53.2 & 48.0 \\     
        MesoInception\cite{afchar2018mesonet} & WIFS'18 & 74.2 & 70.2 & 68.3 \\
        Xception\cite{rossler2019faceforensics++} & ICCV'19  & 81.0 & 81.6 & 79.1 \\
        EfficientNet-B4\cite{tan2019efficientnet} & ICML'19 & 81.5 & 80.8 & 73.8 \\ 
        Capsule\cite{nguyen2019capsule} & ICASSP'19 & 82.9 & 83.4 & 75.4 \\
        F3Net\cite{qian2020thinking} & ECCV'20 & 81.1 & 78.9 & 73.8 \\
        CNN-Aug\cite{wang2020cnn} & CVPR'20 & 79.2 & 74.2 & 66.2 \\
        FFD\cite{dang2020detection} & CVPR'20 & 76.1 & 74.2 & 70.9 \\
        SPSL\cite{liu2021spatial} & CVPR'21 & 85.0 & 79.9 & 74.4 \\
        SRM\cite{luo2021generalizing} & CVPR'21 & 83.6 & 84.0 & 79.4 \\
        RFM\cite{wang2021representative} & CVPR'21 & 85.5 & 82.6 & 74.4 \\
        MATT\cite{zhao2021multi} & CVPR'21 & 79.2 & 76.0 & 74.8 \\
        CLIP\cite{radford2021learning} & ICML'21 & 89.4 & 88.2 & 79.2 \\
        RECCE\cite{cao2022end} & CVPR'22 & 81.5 & 82.3 & 80.8 \\
        SBI\cite{shiohara2022detecting} & CVPR'22 & 71.2 & 79.4 & 73.4 \\
        CORE\cite{ni2022core} & CVPRW'22 & 76.5 & 80.9 & 74.9 \\ 
        SIA\cite{sun2022information} & ECCV'22 & 86.3 & 79.1 & 70.0 \\
        UCF\cite{yan2023ucf} & ICCV'23 & 86.1 & 83.7 & 76.1 \\
        IID\cite{huang2023implicit} & CVPR'23 & 77.2 & 80.7 & 75.6 \\
        LSDA\cite{yan2024transcending} & CVPR'24 & 79.2 & 77.7 & 72.7 \\
        CFM\cite{luo2023beyond} & TIFS'24 & 88.9 & 87.5 & 76.5 \\
        ProDet\cite{cheng2024can} & NeurIPS'24 & 94.5 & 92.6 & 73.6 \\
        ForAda\cite{cui2024forensics} & CVPR'25 & 96.9 & 95.7 & 75.1 \\
        Effort\cite{yan2024effort} & ICML'25 & 92.7 & 93.8 & 85.1 \\
    \hline 
    Average & - & 81.6 & 80.9 & 73.8 \\
    \hline
    \end{tabular}
\end{table}

\subsection{Highlighting the Increased Challenge} 
\label{sec:challenge}
To highlight the challenge, we compare Celeb-DF++ with our previous Celeb-DF datasets using all 24 detectors under a widely used evaluation protocol, following previous works~\cite{zhao2021multi,yan2023ucf,cui2024forensics}. In this setting, detectors are trained using the FF++ (HQ) dataset~\cite{rossler2019faceforensics++} and directly tested on the others. For comprehensiveness, we report both frame-level and video-level ROC AUC scores. 

Table \ref{tab:frame3} and \ref{tab:video3} show the frame-level and video-level evaluation results of both the preview (v1) and official (v2) versions of Celeb-DF~\cite{li2020celeb} and Celeb-DF++. Note that Celeb-DF++ contains a more diverse range of DeepFake methods than Celeb-DF. To evaluate Celeb-DF++, for a specific detection method, we calculate the AUC score of each DeepFake method using the same set of real videos and corresponding DeepFake videos, and then average these scores to obtain the final AUC performance. 
The results show that all DeepFake detectors notably drop their performance on Celeb-DF++, with an average decrease of approximately $5.2\%$ in frame-level AUC and $7.1\%$ in video-level AUC, compared to Celeb-DF. These findings highlight the increased challenge of the Celeb-DF++ dataset. 

\subsection{Generalizable Evaluations}
To comprehensively evaluate the generalizability of DeepFake detection methods, we describe three evaluation protocols: Generalized Forgery evaluation (GF-eval), Generalized Forgery across Quality evaluation (GFQ-eval), and Generalized Forgery across Datasets evaluation (GFD-eval), respectively. 

\begin{table*}[!t]
    \centering
    \caption{\small \textbf{Protocol \#1 (GF-eval): Frame-level AUC ($\%$) results.} All detectors are trained using Celeb-DF and tested on other DeepFake methods in Celeb-DF++. The top-1 and top-2 performance are highlighted by \textbf{bold} and \uline{underscore}.}
    \vspace{-0.3cm}
    \setlength{\tabcolsep}{4pt}
    \label{tab:cross_method}
    \resizebox{1.0\linewidth}{!}{
    \begin{tabular}{l|c|c|c|c|c|c|c|c|c|c|c|c|c|c|c|c|c|c|c|c|c|>{\columncolor{hl}}c}
    \hline
        
            & \multicolumn{7}{c|}{Face-swap (FS)}
            & \multicolumn{7}{c|}{Face-reenactment (FR)}
            & \multicolumn{7}{c|}{Talking-face (TF)}
            & \\
                    \cline{2-22}
    Detector & \rotatebox{90}{BlendFace\cite{shiohara2023blendface}} & \rotatebox{90}{GHOST\cite{groshev2022ghost}} & \rotatebox{90}{HifiFace\cite{wang2021hififace}} & \rotatebox{90}{InSwapper\cite{inswapper}} & \rotatebox{90}{MobileFaceSwap\cite{xu2022mobilefaceswap}} & \rotatebox{90}{SimSwap\cite{chen2020simswap}} & \rotatebox{90}{UniFace\cite{xu2022designing}} & \rotatebox{90}{DaGAN\cite{hong2022depth}} & \rotatebox{90}{FSRT\cite{rochow2024fsrt}} & \rotatebox{90}{HyperReenact\cite{bounareli2023hyperreenact}} & \rotatebox{90}{LIA\cite{wang2024lia}} & \rotatebox{90}{LivePortrait\cite{guo2024liveportrait}} & \rotatebox{90}{MCNET\cite{hong2023implicit}} & \rotatebox{90}{TPSMM\cite{zhao2022thin}} & \rotatebox{90}{AniTalker\cite{liu2024anitalker}} & \rotatebox{90}{EchoMimic\cite{chen2025echomimic}} & \rotatebox{90}{EDTalk\cite{tan2024edtalk}} & \rotatebox{90}{FLOAT\cite{ki2024float}} & \rotatebox{90}{IP-LAP\cite{zhong2023identity}} & \rotatebox{90}{Real3DPortrait\cite{ye2024real3d}} & \rotatebox{90}{SadTalker\cite{zhang2023sadtalker}} & \rotatebox{90}{Average}\\
    
    \hline
        Xception\cite{rossler2019faceforensics++} & 82.6 & 56.4 & 70.5 & 70.9 & 81.7 & 58.6 & 64.2 & 70.0 & 86.1 & 89.8 & 77.7 & 65.9 & 78.4 & 79.5 & 76.4 & 54.2 & 76.7 & 79.2 & 62.7 & 79.3 & 56.6 & \uline{72.3}\\
        RFM\cite{wang2021representative} & 84.0 & 57.0 & 71.7 & 70.6 & 84.4 & 62.0 & 63.5 & 67.6 & 81.2 & 90.6 & 74.2 & 67.2 & 77.2 & 77.6 &71.6 &48.8 & 71.0 & 76.7 & 65.2 & 73.3 & 54.8 & 71.0 \\
        CLIP\cite{radford2021learning} & 84.5 & 74.3 & 72.8 & 77.2 & 86.8 & 71.5 & 66.2 & 65.5 & 76.7 & 80.2 & 72.1 & 54.8 & 70.6 & 72.2 & 62.7 & 46.2 & 67.4 & 62.5 & 63.9 & 70.2 & 53.7 & 69.1 \\
        SIA\cite{sun2022information} & 82.2 & 58.2 & 70.8 & 70.5 & 85.8 & 63.5 & 63.1 & 68.7 & 78.9 & 86.5 & 71.1 & 67.9 & 75.5 & 76.4 & 70.5 & 50.5 & 71.1 & 71.4 & 62.9 & 74.8 & 56.4 & 70.3 \\
        UCF\cite{yan2023ucf} & 79.8 & 59.2 & 63.3 & 64.9 & 85.0 & 56.9 & 60.3 & 63.6 & 79.8 & 88.3 & 69.2 & 62.9 & 72.2 & 73.2 & 68.4 & 44.9 & 65.6 & 71.5 & 65.6 & 73.2 & 50.2 & {67.5} \\
        IID\cite{huang2023implicit} & 80.9 & 51.7 & 70.2 & 67.6 & 84.1 & 54.8 & 61.2 & 67.9 & 82.6 & 91.4 & 75.6 & 59.6 & 73.1 & 74.1 & 72.4 & 49.3 & 69.1 & 75.4 & 58.6 & 76.9 & 46.7 & 68.7 \\
        ProDet\cite{cheng2024can} & 82.1 & 57.9 & 76.8 & 74.6 & 83.5 & 65.1 & 70.0 & 66.9 & 79.7 & 90.0 & 74.5 & 68.5 & 75.9 & 75.9 & 64.9 & 56.4 & 61.7 & 72.3 & 71.3 & 77.8 & 54.5 & 71.4 \\
        Effort\cite{yan2024effort} & 97.9 & 92.2 & 95.2 & 95.4 & 96.8 & 90.2 & 97.0 & 76.6 & 87.2 & 96.9 & 80.9 & 69.7 & 80.1 & 80.2 & 74.6 & 59.6 & 83.5 & 83.1 & 65.2 & 84.1 & 57.5 & \textbf{83.0} \\
        \hline
        Average & 84.3 & 63.4 & 73.9 & 73.0 & {86.0} & 65.3 & 68.2 & 68.4 & 81.5 & {89.2} & 74.4 & 64.6 & 75.4 & 76.1 & 70.2 & {51.2} & 70.8 & 74.0 & 64.4 & 76.2 & 53.8 & 71.7\\
    \hline 
    
    \end{tabular}}
\end{table*}
\begin{table*}[!t]
    \centering
    \caption{\small \textbf{Protocol \#1 (GF-eval): Video-level AUC ($\%$) results.}  All detectors are trained using Celeb-DF and tested on other DeepFake methods in Celeb-DF++. The top-1 and top-2 performance are highlighted by \textbf{bold} and \uline{underscore}.}
    \vspace{-0.3cm}
    \setlength{\tabcolsep}{4pt}
    \label{tab:cross_method_video}
    \resizebox{1.0\linewidth}{!}{
    \begin{tabular}{l|c|c|c|c|c|c|c|c|c|c|c|c|c|c|c|c|c|c|c|c|c|>{\columncolor{hl}}c}
    \hline
        
            & \multicolumn{7}{c|}{Face-swap (FS)}
            & \multicolumn{7}{c|}{Face-reenactment (FR)}
            & \multicolumn{7}{c|}{Talking-face (TF)}
            & \\
                    \cline{2-22}
    Detector & \rotatebox{90}{BlendFace\cite{shiohara2023blendface}} & \rotatebox{90}{GHOST\cite{groshev2022ghost}} & \rotatebox{90}{HifiFace\cite{wang2021hififace}} & \rotatebox{90}{InSwapper\cite{inswapper}} & \rotatebox{90}{MobileFaceSwap\cite{xu2022mobilefaceswap}} & \rotatebox{90}{SimSwap\cite{chen2020simswap}} & \rotatebox{90}{UniFace\cite{xu2022designing}} & \rotatebox{90}{DaGAN\cite{hong2022depth}} & \rotatebox{90}{FSRT\cite{rochow2024fsrt}} & \rotatebox{90}{HyperReenact\cite{bounareli2023hyperreenact}} & \rotatebox{90}{LIA\cite{wang2024lia}} & \rotatebox{90}{LivePortrait\cite{guo2024liveportrait}} & \rotatebox{90}{MCNET\cite{hong2023implicit}} & \rotatebox{90}{TPSMM\cite{zhao2022thin}} & \rotatebox{90}{AniTalker\cite{liu2024anitalker}} & \rotatebox{90}{EchoMimic\cite{chen2025echomimic}} & \rotatebox{90}{EDTalk\cite{tan2024edtalk}} & \rotatebox{90}{FLOAT\cite{ki2024float}} & \rotatebox{90}{IP-LAP\cite{zhong2023identity}} & \rotatebox{90}{Real3DPortrait\cite{ye2024real3d}} & \rotatebox{90}{SadTalker\cite{zhang2023sadtalker}} & \rotatebox{90}{Average}\\
    
    \hline
        Xception\cite{rossler2019faceforensics++} & 86.2 & 52.8 & 70.7 & 70.8 & 85.7 & 56.9 & 66.3 & 70.9 & 90.0 & 92.2 & 81.5 & 66.4 & 82.1 & 83.0 & 75.4 & 47.5 & 75.9 & 79.6 & 62.3 & 75.6 & 46.9 & 72.3\\
        RFM\cite{wang2021representative} & 86.5 & 54.8 & 71.0 & 71.5 & 86.9 & 62.2 & 64.6 & 65.9 & 83.1 & 91.9 & 75.6 & 66.6 & 78.7 & 78.5 &67.7 &43.3 & 68.0 & 75.3 & 63.2 & 67.5 & 45.5 &  69.9\\
        CLIP\cite{radford2021learning} & 87.3 & 75.9 & 74.0 & 79.0 & 89.0 & 72.4 & 68.4 & 63.4 & 77.1 & 80.3 & 71.5 & 52.6 & 69.4 & 70.7 & 57.3 & 39.7 & 63.9 & 60.7 & 61.0 & 63.8 & 45.6 &  {67.8}\\
        SIA\cite{sun2022information} & 85.5 & 57.6 & 72.0 & 71.3 & 89.1 & 62.8 & 63.7 & 69.0 & 80.7 & 88.6 & 73.1 & 68.2 & 77.1 & 78.0 & 69.2 & 44.9 & 70.5 & 71.1 & 61.9 & 72.0 & 48.9 & 70.2 \\
        UCF\cite{yan2023ucf} & 85.4 & 56.9 & 62.6 & 67.2 & 90.2 & 56.5 & 62.1 & 67.3 & 86.2 & 91.6 & 74.7 & 61.6 & 78.5 & 78.9 & 66.0 & 38.5 & 66.6 & 74.2 & 66.0 & 67.8 & 39.6 & 68.5 \\
        IID\cite{huang2023implicit} & 84.9 & 53.1 & 73.5 & 70.6 & 88.5 & 56.5 & 63.1 & 69.7 & 86.2 & 95.1 & 78.4 & 60.8 & 76.2 & 77.4 & 73.6 & 48.2 & 70.3 & 77.8 & 58.2 & 77.2 & 41.2 & 70.5 \\
        ProDet\cite{cheng2024can} & 88.5 & 58.7 & 80.2 & 78.4 & 87.7 & 66.0 & 73.8 & 68.0 & 84.9 & 93.2 & 78.7 & 71.1 & 79.1 & 79.0 & 63.4 & 53.8 & 61.7 & 73.9 & 71.6 & 76.4 & 49.1 & \uline{73.2} \\
        Effort\cite{yan2024effort} & 99.0 & 93.7 & 96.7 & 97.1 & 98.2 & 92.1 & 98.2 & 81.8 & 90.7 & 97.6 & 86.1 & 76.7 & 85.5 & 85.2 & 73.0 & 54.6 & 83.9 & 86.8 & 67.5 & 79.3 & 48.8 & \textbf{84.4} \\
        \hline
        Average & 87.9 & 62.9 & 75.1 & 75.7 & {89.4} & 65.7 & 70.0 & 69.5 & 84.9 & {91.3} & 77.5 & 65.5 & 78.3 & 78.8 & 68.2 & 46.3 & 70.1 & 74.9 & 64.0 & 72.5 & {45.7} & 72.1 \\
    \hline 
    
    \end{tabular}}
\end{table*}

\begin{table*}[!t]
    \centering
    \caption{\small \textbf{Protocol \#2 (GFQ-eval): Frame-level AUC ($\%$) results.} All detectors are trained using Celeb-DF and tested on other DeepFake methods in Celeb-DF++ with compression of \textbf{c35}. The top-1 and top-2 performance are highlighted by \textbf{bold} and \uline{underscore}.}
    \vspace{-0.3cm}
    \setlength{\tabcolsep}{4pt}
    \label{tab:cross_quality_c35}
    \resizebox{1.0\linewidth}{!}{
    \begin{tabular}{l|c|c|c|c|c|c|c|c|c|c|c|c|c|c|c|c|c|c|c|c|c|>{\columncolor{hl}}c}
    \hline

            & \multicolumn{7}{c|}{Face-swap (FS)}
            & \multicolumn{7}{c|}{Face-reenactment (FR)}
            & \multicolumn{7}{c|}{Talking-face (TF)}
            & \\
                    \cline{2-22}
    Detector & \rotatebox{90}{BlendFace\cite{shiohara2023blendface}} & \rotatebox{90}{GHOST\cite{groshev2022ghost}} & \rotatebox{90}{HifiFace\cite{wang2021hififace}} & \rotatebox{90}{InSwapper\cite{inswapper}} & \rotatebox{90}{MobileFaceSwap\cite{xu2022mobilefaceswap}} & \rotatebox{90}{SimSwap\cite{chen2020simswap}} & \rotatebox{90}{UniFace\cite{xu2022designing}} & \rotatebox{90}{DaGAN\cite{hong2022depth}} & \rotatebox{90}{FSRT\cite{rochow2024fsrt}} & \rotatebox{90}{HyperReenact\cite{bounareli2023hyperreenact}} & \rotatebox{90}{LIA\cite{wang2024lia}} & \rotatebox{90}{LivePortrait\cite{guo2024liveportrait}} & \rotatebox{90}{MCNET\cite{hong2023implicit}} & \rotatebox{90}{TPSMM\cite{zhao2022thin}} & \rotatebox{90}{AniTalker\cite{liu2024anitalker}} & \rotatebox{90}{EchoMimic\cite{chen2025echomimic}} & \rotatebox{90}{EDTalk\cite{tan2024edtalk}} & \rotatebox{90}{FLOAT\cite{ki2024float}} & \rotatebox{90}{IP-LAP\cite{zhong2023identity}} & \rotatebox{90}{Real3DPortrait\cite{ye2024real3d}} & \rotatebox{90}{SadTalker\cite{zhang2023sadtalker}} & \rotatebox{90}{Average}\\
    \hline
        Xception\cite{rossler2019faceforensics++} & 76.8 & 62.5 & 64.1 & 67.4 & 76.3 & 62.0 & 64.6 & 70.8 & 81.8 & 68.5 & 73.0 & 68.8 & 75.7 & 76.1 & 72.8 & 59.5 & 68.7 & 75.5 & 62.6 & 80.9 & 63.5 & \uline{70.1} \\
        RFM\cite{wang2021representative} & 77.4 & 65.0 & 65.1 & 67.3 & 80.4 & 66.1 & 64.3 & 68.7 & 77.6 & 73.9 & 70.1 & 69.5 & 74.5 & 74.2 & 68.7 & 52.0 & 59.0 & 72.9 & 61.6 & 75.5 & 60.5 & 68.8 \\
        CLIP\cite{radford2021learning} & 81.6 & 73.3 & 67.2 & 72.2 & 83.9 & 67.0 & 67.9 & 66.9 & 75.9 & 75.3 & 69.8 & 56.8 & 70.6 & 71.2 & 63.1 & 44.2 & 64.8 & 60.1 & 60.6 & 69.3 & 56.2 & 67.5 \\
        SIA\cite{sun2022information} & 75.4 & 61.5 & 64.2 & 67.5 & 80.0 & 65.7 & 66.0 & 67.8 & 75.1 & 72.8 & 67.8 & 70.0 & 72.4 & 72.2 & 66.0 & 56.2 & 58.5 & 67.0 & 60.7 & 76.8 & 60.7 & 67.8 \\
        UCF\cite{yan2023ucf} & 70.7 & 58.8 & 57.4 & 60.1 & 75.4 & 57.2 & 59.1 & 58.2 & 71.3 & 69.2 & 60.0 & 63.3 & 64.2 & 65.2 & 61.4 & 44.1 & 52.2 & 65.6 & 56.6 & 70.9 & 50.2 & {61.5} \\
        IID\cite{huang2023implicit} & 75.5 & 57.5 & 61.0 & 63.5 & 76.1 & 56.5 & 60.5 & 65.5 & 77.3 & 67.2 & 70.0 & 61.0 & 68.0 & 68.5 & 63.4 & 45.4 & 52.2 & 64.8 & 54.9 & 74.3 & 50.4 & 63.5 \\
        ProDet\cite{cheng2024can} & 77.5 & 60.6 & 66.1 & 68.3 & 79.6 & 66.3 & 67.1 & 64.8 & 74.4 & 80.5 & 67.6 & 65.7 & 70.5 & 70.3 & 62.7 & 53.6 & 55.9 & 68.2 & 64.4 & 75.1 & 53.5 & 67.3 \\
        Effort\cite{yan2024effort} & 95.4 & 90.6 & 89.8 & 90.0 & 94.0 & 87.3 & 92.6 & 75.6 & 83.1 & 92.1 & 74.9 & 68.6 & 77.1 & 76.3 & 71.8 & 53.2 & 75.2 & 75.3 & 61.0 & 77.1 & 57.1 & \textbf{79.0} \\
        \hline
        Average & {78.8} & 66.2 & 66.9 & 69.5 & {80.7} & 66.0 & 67.8 & 67.3 & 77.1 & 74.9 & 69.2 & 65.5 & 71.6 & 71.8 & 66.2 & {51.0} & 60.8 & 68.7 & 60.3& 75.0 & 56.5 & 68.2\\
    \hline 
    
    \end{tabular}}
\end{table*}
\begin{table*}[!t]
    \centering
    \caption{\small \textbf{Protocol \#2 (GFQ-eval): Video-level AUC ($\%$) results.} All detectors are trained using Celeb-DF and tested on other DeepFake methods in Celeb-DF++ with compression of \textbf{c35}. The top-1 and top-2 performance are highlighted by \textbf{bold} and \uline{underscore}.}
    \vspace{-0.3cm}
    \setlength{\tabcolsep}{4pt}
    \label{tab:cross_quality_c35_video}
    \resizebox{1.0\linewidth}{!}{
    \begin{tabular}{l|c|c|c|c|c|c|c|c|c|c|c|c|c|c|c|c|c|c|c|c|c|>{\columncolor{hl}}c}
    \hline

            & \multicolumn{7}{c|}{Face-swap (FS)}
            & \multicolumn{7}{c|}{Face-reenactment (FR)}
            & \multicolumn{7}{c|}{Talking-face (TF)}
            & \\
                    \cline{2-22}
    Detector & \rotatebox{90}{BlendFace\cite{shiohara2023blendface}} & \rotatebox{90}{GHOST\cite{groshev2022ghost}} & \rotatebox{90}{HifiFace\cite{wang2021hififace}} & \rotatebox{90}{InSwapper\cite{inswapper}} & \rotatebox{90}{MobileFaceSwap\cite{xu2022mobilefaceswap}} & \rotatebox{90}{SimSwap\cite{chen2020simswap}} & \rotatebox{90}{UniFace\cite{xu2022designing}} & \rotatebox{90}{DaGAN\cite{hong2022depth}} & \rotatebox{90}{FSRT\cite{rochow2024fsrt}} & \rotatebox{90}{HyperReenact\cite{bounareli2023hyperreenact}} & \rotatebox{90}{LIA\cite{wang2024lia}} & \rotatebox{90}{LivePortrait\cite{guo2024liveportrait}} & \rotatebox{90}{MCNET\cite{hong2023implicit}} & \rotatebox{90}{TPSMM\cite{zhao2022thin}} & \rotatebox{90}{AniTalker\cite{liu2024anitalker}} & \rotatebox{90}{EchoMimic\cite{chen2025echomimic}} & \rotatebox{90}{EDTalk\cite{tan2024edtalk}} & \rotatebox{90}{FLOAT\cite{ki2024float}} & \rotatebox{90}{IP-LAP\cite{zhong2023identity}} & \rotatebox{90}{Real3DPortrait\cite{ye2024real3d}} & \rotatebox{90}{SadTalker\cite{zhang2023sadtalker}} & \rotatebox{90}{Average}\\
    \hline
        Xception\cite{rossler2019faceforensics++} & 82.5 & 63.1 & 65.1 & 70.7 & 78.8 & 58.5 & 67.4 & 74.6 & 83.4 & 67.7 & 76.0 & 71.7 & 76.7 & 80.0 & 72.0 & 54.1 & 74.2 & 76.7 & 61.6 & 78.9 & 57.0 & 71.0 \\
        RFM\cite{wang2021representative} & 83.0 & 68.0 & 66.2 & 70.3 & 81.9 & 66.7 & 66.7 & 69.3 & 80.4 & 72.3 & 71.4 & 70.4 & 75.0 & 77.2 & 63.5 & 47.1 & 62.1 & 74.8 & 60.6 & 73.0 & 53.8 & 69.2 \\
        CLIP\cite{radford2021learning} & 86.5 & 77.1 & 69.8 & 75.0 & 84.3 & 65.7 & 70.7 & 65.2 & 74.9 & 74.3 & 68.5 & 54.2 & 67.7 & 69.7 & 56.7 & 36.9 & 64.7 & 59.3 & 56.7 & 62.0 & 46.9 & {66.0} \\
        SIA\cite{sun2022information} & 81.1 & 62.2 & 66.0 & 70.1 & 80.9 & 64.0 & 66.8 & 69.8 & 76.9 & 69.7 & 68.6 & 69.6 & 70.8 & 74.0 & 64.6 & 51.4 & 60.9 & 69.2 & 59.1 & 73.2 & 53.3 & 67.7 \\
        UCF\cite{yan2023ucf} & 81.9 & 64.8 & 60.6 & 68.5 & 82.6 & 57.4 & 66.5 & 70.2 & 82.7 & 73.1 & 70.4 & 67.0 & 73.6 & 78.0 & 66.6 & 44.5 & 70.3 & 75.8 & 60.3 & 72.5 & 46.9 & 68.3 \\
        IID\cite{huang2023implicit} & 81.2 & 61.9 & 63.9 & 67.1 & 80.3 & 58.3 & 64.3 & 69.5 & 80.4 & 70.4 & 73.3 & 63.9 & 68.9 & 72.4 & 64.2 & 45.6 & 55.4 & 71.3 & 53.8 & 74.9 & 45.5 & {66.0} \\
        ProDet\cite{cheng2024can} & 87.1 & 65.7 & 71.9 & 74.2 & 84.6 & 70.1 & 72.9 & 69.9 & 82.1 & 83.6 & 74.6 & 68.9 & 75.1 & 77.3 & 60.5 & 52.7 & 60.0 & 71.6 & 66.0 & 78.1 & 51.5 & \uline{71.4} \\
        Effort\cite{yan2024effort} & 97.6 & 92.3 & 90.8 & 91.8 & 93.6 & 86.4 & 92.7 & 79.2 & 84.1 & 92.9 & 78.6 & 72.8 & 78.9 & 80.1 & 69.9 & 45.9 & 75.6 & 78.5 & 61.6 & 71.4 & 48.2 & \textbf{79.2} \\
        \hline
        Average & {85.1} & 69.4 & 69.3 & 73.5 & {83.4} & 65.9 & 71.0 & 71.0 & 80.6 & 75.5 & 72.7 & 67.3 & 73.3 & 76.1 & 64.8 & {47.3} & 65.4 & 72.2 & 60.0 & 73.0 & 50.4 & 69.9 \\
    \hline 
    
    \end{tabular}}
\end{table*}

\begin{table*}[!ht]
    \centering
    \caption{\small \textbf{Protocol \#2 (GFQ-eval): Frame-level AUC ($\%$) results.} All detectors are trained using Celeb-DF and tested on other DeepFake methods in Celeb-DF++ with compression of \textbf{c45}. The top-1 and top-2 performance are highlighted by \textbf{bold} and \uline{underscore}.}
    \vspace{-0.3cm}
    \setlength{\tabcolsep}{4pt}
    \label{tab:cross_quality_c45}
    \resizebox{1.0\linewidth}{!}{
    \begin{tabular}{l|c|c|c|c|c|c|c|c|c|c|c|c|c|c|c|c|c|c|c|c|c|>{\columncolor{hl}}c}
    \hline

            & \multicolumn{7}{c|}{Face-swap (FS)}
            & \multicolumn{7}{c|}{Face-reenactment (FR)}
            & \multicolumn{7}{c|}{Talking-face (TF)}
            & \\
                    \cline{2-22}
    Detector & \rotatebox{90}{BlendFace\cite{shiohara2023blendface}} & \rotatebox{90}{GHOST\cite{groshev2022ghost}} & \rotatebox{90}{HifiFace\cite{wang2021hififace}} & \rotatebox{90}{InSwapper\cite{inswapper}} & \rotatebox{90}{MobileFaceSwap\cite{xu2022mobilefaceswap}} & \rotatebox{90}{SimSwap\cite{chen2020simswap}} & \rotatebox{90}{UniFace\cite{xu2022designing}} & \rotatebox{90}{DaGAN\cite{hong2022depth}} & \rotatebox{90}{FSRT\cite{rochow2024fsrt}} & \rotatebox{90}{HyperReenact\cite{bounareli2023hyperreenact}} & \rotatebox{90}{LIA\cite{wang2024lia}} & \rotatebox{90}{LivePortrait\cite{guo2024liveportrait}} & \rotatebox{90}{MCNET\cite{hong2023implicit}} & \rotatebox{90}{TPSMM\cite{zhao2022thin}} & \rotatebox{90}{AniTalker\cite{liu2024anitalker}} & \rotatebox{90}{EchoMimic\cite{chen2025echomimic}} & \rotatebox{90}{EDTalk\cite{tan2024edtalk}} & \rotatebox{90}{FLOAT\cite{ki2024float}} & \rotatebox{90}{IP-LAP\cite{zhong2023identity}} & \rotatebox{90}{Real3DPortrait\cite{ye2024real3d}} & \rotatebox{90}{SadTalker\cite{zhang2023sadtalker}} & \rotatebox{90}{Average}\\
    \hline
        Xception\cite{rossler2019faceforensics++} & 68.6 & 61.5 & 64.2 & 65.5 & 71.5 & 64.0 & 60.9 & 69.1 & 72.1 & 56.8 & 66.7 & 70.9 & 70.0 & 73.6 & 56.2 & 48.8 & 63.9 & 59.8 & 64.8 & 72.2 & 60.3 & 64.8 \\
        RFM\cite{wang2021representative} & 68.3 & 63.6 & 62.5 & 64.5 & 74.4 & 65.6 & 63.6 & 69.5 & 71.6 & 60.9 & 67.9 & 72.0 & 70.5 & 73.4 & 60.9 & 44.9 & 55.6 & 58.2 & 60.4 & 68.8 & 61.9 & 64.7 \\
        CLIP\cite{radford2021learning} & 74.9 & 69.6 & 62.7 & 68.7 & 77.7 & 60.5 & 66.5 & 71.2 & 74.7 & 67.5 & 72.3 & 62.7 & 72.7 & 73.3 & 59.9 & 42.2 & 59.5 & 49.7 & 62.8 & 65.5 & 61.3 & \uline{65.5} \\
        SIA\cite{sun2022information} & 68.3 & 60.3 & 58.5 & 63.0 & 71.6 & 64.7 & 63.3 & 64.5 & 66.6 & 62.6 & 61.2 & 71.6 & 69.7 & 70.4 & 59.0 & 48.3 & 54.1 & 53.9 & 59.6 & 71.5 & 60.3 & 63.0 \\
        UCF\cite{yan2023ucf} & 64.8 & 55.9 & 55.9 & 59.0 & 66.5 & 56.7 & 52.4 & 59.3 & 59.9 & 53.7 & 58.0 & 64.2 & 57.2 & 63.2 & 53.3 & 43.6 & 48.2 & 53.4 & 54.4 & 64.5 & 49.3 & {56.8} \\
        IID\cite{huang2023implicit} & 70.5 & 61.9 & 61.6 & 65.3 & 73.8 & 62.3 & 58.1 & 66.2 & 70.7 & 54.8 & 68.0 & 65.5 & 63.9 & 70.2 & 55.5 & 41.0 & 43.9 & 51.2 & 57.5 & 68.7 & 53.2 & 61.1 \\
        ProDet\cite{cheng2024can} & 69.5 & 58.2 & 61.7 & 64.4 & 73.4 & 65.9 & 61.5 & 66.8 & 68.5 & 66.7 & 65.7 & 67.8 & 70.2 & 68.9 & 61.0 & 45.0 & 56.8 & 59.6 & 68.0 & 65.8 & 58.6 & 64.0 \\
        Effort\cite{yan2024effort} & 84.3 & 83.3 & 81.3 & 81.8 & 85.4 & 79.3 & 84.2 & 66.7 & 69.5 & 80.4 & 63.8 & 67.8 & 69.0 & 71.8 & 64.1 & 41.9 & 64.2 & 61.5 & 55.0 & 62.4 & 55.3 & \textbf{70.1} \\
        \hline
        Average & {71.1} & 64.3 & 63.6 & 66.5 & {74.3} & 64.9 & 63.8 & 66.7 & 69.2 & 62.9 & 65.5 & 67.8 & 67.9 & 70.6 & 58.7 & {44.5} & 55.8 & 55.9 & 60.3 &67.4 & 57.5& 63.8\\
    \hline 
    
    \end{tabular}}
\end{table*}
\begin{table*}[!ht]
    \centering
    \caption{\small \textbf{Protocol \#2 (GFQ-eval): Video-level AUC ($\%$) results.} All detectors are trained using Celeb-DF and tested on other DeepFake methods in Celeb-DF++ with compression of \textbf{c45}. The top-1 and top-2 performance are highlighted by \textbf{bold} and \uline{underscore}.}
    \vspace{-0.3cm}
    \setlength{\tabcolsep}{4pt}
    \label{tab:cross_quality_c45_video}
    \resizebox{1.0\linewidth}{!}{
    \begin{tabular}{l|c|c|c|c|c|c|c|c|c|c|c|c|c|c|c|c|c|c|c|c|c|>{\columncolor{hl}}c}
    \hline

            & \multicolumn{7}{c|}{Face-swap (FS)}
            & \multicolumn{7}{c|}{Face-reenactment (FR)}
            & \multicolumn{7}{c|}{Talking-face (TF)}
            & \\
                    \cline{2-22}
    Detector & \rotatebox{90}{BlendFace\cite{shiohara2023blendface}} & \rotatebox{90}{GHOST\cite{groshev2022ghost}} & \rotatebox{90}{HifiFace\cite{wang2021hififace}} & \rotatebox{90}{InSwapper\cite{inswapper}} & \rotatebox{90}{MobileFaceSwap\cite{xu2022mobilefaceswap}} & \rotatebox{90}{SimSwap\cite{chen2020simswap}} & \rotatebox{90}{UniFace\cite{xu2022designing}} & \rotatebox{90}{DaGAN\cite{hong2022depth}} & \rotatebox{90}{FSRT\cite{rochow2024fsrt}} & \rotatebox{90}{HyperReenact\cite{bounareli2023hyperreenact}} & \rotatebox{90}{LIA\cite{wang2024lia}} & \rotatebox{90}{LivePortrait\cite{guo2024liveportrait}} & \rotatebox{90}{MCNET\cite{hong2023implicit}} & \rotatebox{90}{TPSMM\cite{zhao2022thin}} & \rotatebox{90}{AniTalker\cite{liu2024anitalker}} & \rotatebox{90}{EchoMimic\cite{chen2025echomimic}} & \rotatebox{90}{EDTalk\cite{tan2024edtalk}} & \rotatebox{90}{FLOAT\cite{ki2024float}} & \rotatebox{90}{IP-LAP\cite{zhong2023identity}} & \rotatebox{90}{Real3DPortrait\cite{ye2024real3d}} & \rotatebox{90}{SadTalker\cite{zhang2023sadtalker}} & \rotatebox{90}{Average}\\
    \hline
        Xception\cite{rossler2019faceforensics++} & 77.6 & 55.0 & 56.6 & 62.6 & 61.6 & 54.4 & 54.9 & 72.1 & 68.7 & 51.0 & 73.0 & 72.8 & 66.1 & 67.6 & 49.9 & 45.6 & 69.7 & 61.4 & 66.3 & 70.8 & 56.4 & 62.6 \\
        RFM\cite{wang2021representative} & 76.3 & 60.1 & 57.5 & 62.0 & 63.2 & 57.6 & 56.4 & 63.3 & 65.7 & 54.4 & 66.3 & 66.8 & 64.5 & 65.7 & 58.0 & 42.7 & 58.6 & 56.7 & 62.4 & 69.8 & 54.8 & 61.1 \\
        CLIP\cite{radford2021learning} & 83.4 & 69.7 & 63.8 & 67.0 & 73.7 & 58.6 & 66.2 & 68.9 & 69.8 & 60.5 & 68.2 & 60.8 & 66.2 & 68.0 & 46.4 & 39.4 & 60.8 & 52.8 & 60.9 & 59.7 & 45.8 & 62.4 \\
        SIA\cite{sun2022information} & 76.0 & 54.2 & 56.5 & 65.3 & 63.8 & 54.5 & 56.7 & 68.8 & 63.5 & 58.4 & 67.1 & 70.5 & 62.8 & 64.8 & 51.4 & 48.8 & 55.3 & 58.2 & 60.5 & 69.8 & 53.3 & 61.0 \\
        UCF\cite{yan2023ucf} & 74.6 & 55.4 & 53.9 & 59.9 & 62.2 & 56.1 & 50.7 & 65.8 & 63.2 & 45.4 & 63.6 & 66.1 & 60.4 & 62.0 & 46.5 & 43.8 & 61.0 & 55.6 & 58.5 & 66.1 & 45.8 & {57.9} \\
        IID\cite{huang2023implicit} & 80.4 & 56.8 & 58.0 & 63.3 & 64.3 & 52.5 & 55.3 & 66.7 & 66.3 & 52.4 & 70.4 & 68.9 & 57.1 & 62.5 & 45.2 & 40.4 & 44.8 & 52.6 & 58.9 & 69.8 & 49.7 & 58.9 \\
        ProDet\cite{cheng2024can} & 80.2 & 57.6 & 61.8 & 63.8 & 68.9 & 61.1 & 61.5 & 69.9 & 68.5 & 70.5 & 69.8 & 69.0 & 67.1 & 67.6 & 51.2 & 45.1 & 61.7 & 59.9 & 69.6 & 74.7 & 56.6 & \uline{64.6} \\
        Effort\cite{yan2024effort} & 91.6 & 82.5 & 79.2 & 84.0 & 82.9 & 74.9 & 83.3 & 70.7 & 67.7 & 82.8 & 68.8 & 70.9 & 69.5 & 68.5 & 60.8 & 42.8 & 64.3 & 66.4 & 59.8 & 64.2 & 48.5 & \textbf{70.7} \\
        \hline
        Average & {80.0} & 61.4 & 60.9 & 66.0 & 67.6 & 58.7 & 60.6 & 68.3 & 66.7 & 59.4 & {68.4} & 68.2 & 64.2 & 65.8 & 51.2 & {43.6} & 59.5 & 58.0 & 62.1 & 68.1 & 51.4 & 62.4 \\

    \hline 
    
    \end{tabular}}
\end{table*}
\begin{table*}[!ht]
    \centering
    \caption{\small \textbf{Protocol \#3 (GFD-eval): Frame-level AUC ($\%$) results.} All detectors are trained on FF++ (HQ) and tested on all DeepFake methods in Celeb-DF++. The top-1 and top-2 performance are highlighted by \textbf{bold} and \uline{underscore}.}
    \vspace{-0.3cm}
    \setlength{\tabcolsep}{3pt}
    \label{tab:22-frame-level}
    \resizebox{1.0\linewidth}{!}{
    \begin{tabular}{l|c|c|c|c|c|c|c|c|c|c|c|c|c|c|c|c|c|c|c|c|c|c|>
    {\columncolor{hl}}c}
    \hline

             & \multicolumn{8}{c|}{Face-swap (FS)}
            & \multicolumn{7}{c|}{Face-reenactment (FR)}
            & \multicolumn{7}{c|}{Talking-face (TF)} \\
                    \cline{2-23}

    Detector & \rotatebox{90}{Celeb-DF\cite{li2020celeb}} & \rotatebox{90}{BlendFace\cite{shiohara2023blendface}} & \rotatebox{90}{GHOST\cite{groshev2022ghost}} & \rotatebox{90}{HifiFace\cite{wang2021hififace}} & \rotatebox{90}{InSwapper\cite{inswapper}} & \rotatebox{90}{MobileFaceSwap\cite{xu2022mobilefaceswap}} & \rotatebox{90}{SimSwap\cite{chen2020simswap}} & \rotatebox{90}{UniFace\cite{xu2022designing}} & \rotatebox{90}{DaGAN\cite{hong2022depth}} & \rotatebox{90}{FSRT\cite{rochow2024fsrt}} & \rotatebox{90}{HyperReenact\cite{bounareli2023hyperreenact}} & \rotatebox{90}{LIA\cite{wang2024lia}} & \rotatebox{90}{LivePortrait\cite{guo2024liveportrait}} & \rotatebox{90}{MCNET\cite{hong2023implicit}} & \rotatebox{90}{TPSMM\cite{zhao2022thin}} & \rotatebox{90}{AniTalker\cite{liu2024anitalker}} & \rotatebox{90}{EchoMimic\cite{chen2025echomimic}} & \rotatebox{90}{EDTalk\cite{tan2024edtalk}} & \rotatebox{90}{FLOAT\cite{ki2024float}} & \rotatebox{90}{IP-LAP\cite{zhong2023identity}} & \rotatebox{90}{Real3DPortrait\cite{ye2024real3d}} & \rotatebox{90}{SadTalker\cite{zhang2023sadtalker}}& \rotatebox{90}{Average}\\ 
    \hline
        MesoNet\cite{afchar2018mesonet} & 53.1 & 57.3 & 44.5 & 46.4 & 50.6 & 46.4 & 45.0 & 48.8 & 41.9 & 46.8 & 44.7 & 56.5 & 38.6 & 42.8 & 43.1 & 46.3 & 51.2 & 64.7 & 41.6 & 51.3 & 49.1 & 43.9 & {47.9}\\     
        MesoInception\cite{afchar2018mesonet} & 65.0 & 62.7 & 56.3 & 63.5 & 67.6 & 61.7 & 60.0 & 63.9 & 63.6 & 67.4 & 58.9 & 62.7 & 56.8 & 68.3 & 68.0 & 55.5 & 63.3 & 64.0 & 54.4 & 80.0 & 76.8 & 63.1 & 63.8\\
        Xception\cite{rossler2019faceforensics++} & 74.0 & 73.8 & 56.0 & 78.3 & 81.6 & 71.1 & 54.3 & 74.4 & 70.2 & 82.2 & 77.1 & 79.3 & 59.0 & 79.8 & 80.9 & 75.5 & 68.7 & 84.4 & 64.2 & 84.6 & 89.1 & 57.9 &73.5\\
        EfficientNet-B4\cite{tan2019efficientnet} & 75.0 & 73.3 & 56.9 & 73.2 & 73.4 & 65.5 & 50.3 & 68.4 & 64.2 & 76.0 & 70.6 & 75.0 & 55.8 & 74.0 & 75.9 & 66.3 & 69.1 & 83.6 & 67.5 & 80.8 & 86.9 & 58.5 &70.0\\ 
        Capsule\cite{nguyen2019capsule} & 76.5 & 67.1 & 52.9 & 72.9 & 72.5 & 66.7 & 57.1 & 69.1 & 67.9 & 79.7 & 74.8 & 72.9 & 61.8 & 76.6 & 77.4 & 63.8 & 68.2 & 72.3 & 61.7 & 86.7 & 90.0 & 63.4 &70.5\\
        F3Net\cite{qian2020thinking} & 72.9 & 71.5 & 51.2 & 73.8 & 76.5 & 67.8 & 51.3 & 68.1 & 66.4 & 80.2 & 64.4 & 80.9 & 56.0 & 78.9 & 80.2 & 69.9 & 67.9 & 83.3 & 55.1 & 83.9 & 82.5 & 56.8 &70.0\\
        CNN-Aug\cite{wang2020cnn} & 67.5 & 66.9 & 57.0 & 67.5 & 65.4 & 64.3 & 62.4 & 64.3 & 58.7 & 68.5 & 59.0 & 61.1 & 53.6 & 61.4 & 62.4 & 52.2 & 58.0 & 54.3 & 53.7 & 71.9 & 69.9 & 59.1 &61.8\\
        FFD\cite{dang2020detection} & 68.7 & 67.2 & 52.7 & 70.7 & 73.2 & 65.8 & 47.1 & 60.4 & 64.6 & 74.4 & 64.4 & 73.3 & 51.7 & 73.2 & 73.7 & 71.0 & 59.8 & 79.7 & 63.0 & 80.6 & 79.0 & 52.4 &66.7\\
        SPSL\cite{liu2021spatial} & 72.9 & 72.1 & 59.8 & 69.5 & 73.4 & 72.2 & 53.9 & 65.6 & 62.3 & 70.4 & 77.7 & 65.9 & 53.5 & 68.3 & 68.2 & 69.2 & 69.0 & 72.1 & 65.3 & 79.1 & 83.1 & 51.7 &68.0\\
        SRM\cite{luo2021generalizing} & 76.0 & 71.7 & 55.1 & 76.9 & 82.4 & 69.5 & 58.4 & 68.8 & 71.6 & 79.5 & 74.8 & 80.4 & 64.7 & 79.0 & 79.5 & 67.7 & 68.8 & 80.1 & 58.2 & 84.6 & 88.1 & 59.6 &72.5\\
        RFM\cite{wang2021representative} & 76.4 & 81.0 & 63.2 & 75.3 & 82.4 & 72.5 & 49.7 & 76.8 & 62.4 & 76.0 & 70.3 & 72.5 & 60.8 & 73.9 & 75.4 & 67.8 & 65.2 & 80.2 & 60.6 & 77.4 & 76.2 & 51.6 &70.3\\
        MATT\cite{zhao2021multi} & 72.0 & 68.2 & 53.1 & 68.8 & 74.4 & 65.6 & 58.1 & 60.1 & 65.5 & 72.7 & 68.6 & 63.2 & 57.1 & 68.8 & 69.9 & 73.7 & 61.5 & 79.5 & 47.2 & 76.3 & 75.8 & 65.2 &66.6\\
        CLIP\cite{radford2021learning} & 82.7 & 74.2 & 65.3 & 75.5 & 81.3 & 74.6 & 70.7 & 69.4 & 75.3 & 84.1 & 68.4 & 81.1 & 56.6 & 80.7 & 82.3 & 74.9 & 71.5 & 79.3 & 55.0 & 89.6 & 81.5 & 73.4 &74.9\\
        RECCE\cite{cao2022end} & 74.1 & 76.9 & 62.3 & 77.2 & 84.4 & 66.7 & 60.0 & 73.1 & 76.3 & 87.4 & 76.6 & 84.3 & 65.9 & 84.9 & 85.4 & 74.7 & 70.2 & 86.7 & 54.1 & 88.7 & 89.6 & 63.8 & \uline{75.6}\\
        SBI\cite{shiohara2022detecting} & 74.8 &85.1&57.0&79.2&	83.5&73.3&59.7&62.5&68.8&75.3&65.6&78.6&65.7&75.6&77.9&53.6&64.2&84.7&48.0&84.0&86.5&62.7&71.2\\
        CORE\cite{ni2022core} & 74.1 & 73.8 & 55.5 & 75.6 & 83.3 & 72.4 & 53.5 & 70.4 & 63.6 & 76.3 & 73.6 & 76.3 & 56.1 & 76.4 & 76.7 & 69.8 & 63.3 & 86.2 & 52.6 & 80.3 & 86.1 & 51.7 &70.3\\ 
        SIA\cite{sun2022information} & 72.5 & 62.2 & 55.1 & 65.3 & 65.7 & 65.9 & 57.2 & 63.4 & 62.1 & 68.4 & 61.6 & 61.2 & 50.1 & 65.5 & 66.8 & 65.7 & 65.1 & 69.0 & 61.0 & 75.9 & 75.5 & 61.2 &64.4\\
        UCF\cite{yan2023ucf} & 77.2 & 75.5 & 57.0 & 76.8 & 79.9 & 73.6 & 55.6 & 70.5 & 66.9 & 81.8 & 72.8 & 80.3 & 56.7 & 79.0 & 79.8 & 73.0 & 65.4 & 84.4 & 63.9 & 81.0 & 85.3 & 52.7 &72.2\\
        IID\cite{huang2023implicit} & 74.7 & 76.7 & 58.2 & 75.2 & 79.5 & 70.8 & 57.1 & 68.8 & 69.3 & 83.2 & 71.4 & 80.8 & 62.3 & 80.8 & 81.8 & 68.5 & 63.8 & 80.1 & 50.5 & 83.8 & 82.1 & 57.2 &71.7\\
        LSDA\cite{yan2024transcending} & 73.7 & 77.4 & 60.1 & 78.3 & 78.0 & 64.2 & 51.0 & 71.3 & 63.3 & 78.7 & 66.2 & 76.0 & 61.7 & 75.3 & 76.4 & 63.9 & 67.1 & 78.2 & 58.2 & 77.8 & 85.2 & 55.8 &69.9\\
        CFM\cite{luo2023beyond} & 81.1 & 80.3 & 64.4 & 76.5 & 78.3 & 69.9 & 57.5 & 77.2 & 66.4 & 79.6 & 66.2 & 80.8 & 63.2 & 81.0 & 82.5 & 69.0 & 71.4 & 84.9 & 50.7 & 85.0 & 84.5 & 60.8 &73.2\\
        ProDet\cite{cheng2024can} & 84.2 & 83.8 & 55.5 & 79.0 & 81.9 & 81.3 & 52.1 & 73.7 & 51.0 & 64.5 & 68.8 & 67.9 & 54.7 & 63.6 & 66.1 & 61.0 & 66.1 & 81.5 & 66.4 & 76.0 & 86.8 & 46.3 &68.7\\
        ForAda\cite{cui2024forensics} & 89.9 & 84.1 & 67.5 & 83.3 & 87.8 & 82.1 & 70.4 & 86.4 & 61.6 & 65.2 & 71.8 & 69.4 & 56.0 & 69.7 & 67.3 & 68.6 & 58.9 & 84.3 & 56.4 & 72.5 & 73.8 & 53.7 &71.9\\
        Effort\cite{yan2024effort} & 86.8 & 84.6 & 67.0 & 80.8 & 83.7 & 77.1 & 67.6 & 89.7 & 75.8 & 83.1 & 86.2 & 83.3 & 59.6 & 84.7 & 83.7 & 81.9 & 78.1 & 93.8 & 78.2 & 84.9 & 92.3 & 65.8 &\textbf{80.4}\\
        \hline
        Average & 74.8 & 73.6 & 57.7 &73.3 &76.7 &69.2 &{56.7} &69.4 & 65.0& 75.1 & 68.9 & 73.5 & 57.4 & 73.4 & 74.2 & 66.8 & 65.7 & 78.8 & 57.8 & {79.9} & {81.5} & 57.8 & 69.4 \\
    \hline 
    
    \end{tabular}}
\end{table*}

\begin{table*}[!ht]
    \centering
    \caption{\small \textbf{Protocol \#3 (GFD-eval): Video-level AUC ($\%$) results.} All detectors are trained on FF++ (HQ) and tested on all DeepFake methods in Celeb-DF++. The top-1 and top-2 performance are highlighted by \textbf{bold} and \uline{underscore}.}
    \vspace{-0.3cm}
    \setlength{\tabcolsep}{3pt}
    \label{tab:22-video-level}
    \resizebox{1.0\linewidth}{!}{
    \begin{tabular}{l|c|c|c|c|c|c|c|c|c|c|c|c|c|c|c|c|c|c|c|c|c|c|>
    {\columncolor{hl}}c}
    \hline

             & \multicolumn{8}{c|}{Face-swap (FS)}
            & \multicolumn{7}{c|}{Face-reenactment (FR)}
            & \multicolumn{7}{c|}{Talking-face (TF)} \\
                    \cline{2-23}

    Detector & \rotatebox{90}{Celeb-DF\cite{li2020celeb}} & \rotatebox{90}{BlendFace\cite{shiohara2023blendface}} & \rotatebox{90}{GHOST\cite{groshev2022ghost}} & \rotatebox{90}{HifiFace\cite{wang2021hififace}} & \rotatebox{90}{InSwapper\cite{inswapper}} & \rotatebox{90}{MobileFaceSwap\cite{xu2022mobilefaceswap}} & \rotatebox{90}{SimSwap\cite{chen2020simswap}} & \rotatebox{90}{UniFace\cite{xu2022designing}} & \rotatebox{90}{DaGAN\cite{hong2022depth}} & \rotatebox{90}{FSRT\cite{rochow2024fsrt}} & \rotatebox{90}{HyperReenact\cite{bounareli2023hyperreenact}} & \rotatebox{90}{LIA\cite{wang2024lia}} & \rotatebox{90}{LivePortrait\cite{guo2024liveportrait}} & \rotatebox{90}{MCNET\cite{hong2023implicit}} & \rotatebox{90}{TPSMM\cite{zhao2022thin}} & \rotatebox{90}{AniTalker\cite{liu2024anitalker}} & \rotatebox{90}{EchoMimic\cite{chen2025echomimic}} & \rotatebox{90}{EDTalk\cite{tan2024edtalk}} & \rotatebox{90}{FLOAT\cite{ki2024float}} & \rotatebox{90}{IP-LAP\cite{zhong2023identity}} & \rotatebox{90}{Real3DPortrait\cite{ye2024real3d}} & \rotatebox{90}{SadTalker\cite{zhang2023sadtalker}}& \rotatebox{90}{Average}\\   
    \hline
        MesoNet\cite{afchar2018mesonet} & 53.2 & 58.0 & 44.3 & 46.2 & 50.7 & 46.2 & 44.7 & 48.8 & 41.3 & 46.3 & 44.0 & 56.5 & 37.4 & 42.0 & 42.4 & 45.7 & 51.0 & 65.1 & 40.9 & 50.8 & 48.4 & 43.3 &{47.6}\\     
        MesoInception\cite{afchar2018mesonet} & 70.2 & 67.0 & 58.7 & 68.0 & 73.3 & 66.1 & 63.5 & 68.4 & 67.8 & 73.0 & 62.4 & 66.7 & 60.2 & 73.8 & 73.5 & 56.9 & 67.2 & 69.0 & 55.3 & 87.0 & 82.3 & 67.4 &68.1\\
        Xception\cite{rossler2019faceforensics++} & 81.6 & 80.1 & 57.5 & 85.0 & 88.1 & 78.8 & 55.2 & 81.0 & 76.3 & 89.3 & 85.3 & 86.7 & 61.6 & 87.4 & 88.4 & 80.5 & 72.6 & 90.6 & 67.4 & 91.8 & 93.8 & 59.4 &79.0\\
        EfficientNet-B4\cite{tan2019efficientnet} & 80.8 & 78.3 & 56.5 & 78.0 & 77.6 & 68.5 & 48.2 & 72.1 & 66.3 & 80.4 & 75.0 & 79.9 & 55.4 & 78.8 & 81.4 & 68.7 & 71.1 & 89.1 & 69.6 & 86.9 & 90.7 & 57.8 &73.2\\ 
        Capsule\cite{nguyen2019capsule} & 83.5 & 71.5 & 52.6 & 78.3 & 76.8 & 71.4 & 59.2 & 73.7 & 72.3 & 84.7 & 81.1 & 78.3 & 65.9 & 82.0 & 83.3 & 66.7 & 71.9 & 77.3 & 64.0 & 93.1 & 95.1 & 66.2 &75.0\\
        F3Net\cite{qian2020thinking} & 78.9 & 76.1 & 47.8 & 79.6 & 81.9 & 72.3 & 50.9 & 71.7 & 71.0 & 86.5 & 68.0 & 87.1 & 56.3 & 85.2 & 86.4 & 73.1 & 70.9 & 88.3 & 54.4 & 91.0 & 85.8 & 55.9 &73.6\\
        CNN-Aug\cite{wang2020cnn} & 74.2 & 72.8 & 61.1 & 73.5 & 71.0 & 70.1 & 67.3 & 69.6 & 62.9 & 75.2 & 64.3 & 65.7 & 56.5 & 66.1 & 67.5 & 53.7 & 60.9 & 57.0 & 55.2 & 78.3 & 75.4 & 62.9 &66.4\\
        FFD\cite{dang2020detection} & 74.2 & 71.8 & 53.0 & 76.0 & 78.2 & 71.1 & 45.4 & 62.6 & 68.5 & 80.6 & 69.0 & 79.7 & 51.8 & 79.4 & 79.9 & 75.3 & 61.1 & 85.5 & 64.6 & 87.5 & 84.1 & 50.5 &70.4\\
        SPSL\cite{liu2021spatial} & 79.9 & 79.0 & 63.6 & 76.2 & 80.6 & 79.8 & 55.3 & 71.2 & 67.4 & 78.2 & 85.7 & 72.3 & 55.1 & 75.9 & 76.0 & 75.1 & 75.7 & 79.8 & 69.5 & 87.6 & 90.2 & 53.4 &74.0\\
        SRM\cite{luo2021generalizing} & 84.0 & 77.9 & 56.6 & 83.7 & 88.5 & 77.9 & 62.6 & 74.4 & 79.8 & 89.9 & 81.6 & 89.2 & 69.2 & 88.5 & 88.7 & 74.8 & 73.2 & 87.5 & 64.1 & 94.0 & 95.1 & 63.4 &79.3\\
        RFM\cite{wang2021representative} & 82.6 & 87.2 & 65.6 & 85.5 & 87.8 & 78.6 & 48.6 & 82.7 & 65.7 & 82.0 & 75.1 & 77.9 & 62.9 & 79.6 & 81.3 & 70.9 & 68.4 & 85.3 & 61.8 & 84.0 & 79.4 & 49.6 &74.7\\
        MATT\cite{zhao2021multi} & 76.0 & 73.9 & 53.3 & 75.4 & 84.2 & 68.3 & 64.2 & 64.5 & 75.9 & 85.0 & 75.2 & 74.7 & 64.7 & 80.9 & 82.7 & 80.7 & 68.3 & 90.4 & 43.4 & 87.7 & 89.4 & 78.2 &74.4\\
        CLIP\cite{radford2021learning} & 88.2 & 79.4 & 68.3 & 80.3 & 86.6 & 80.1 & 75.6 & 73.6 & 80.1 & 89.7 & 73.1 & 86.0 & 56.9 & 85.9 & 87.4 & 79.2 & 74.5 & 84.3 & 55.4 & 94.6 & 84.6 & 76.0 &79.1\\
        RECCE\cite{cao2022end} & 82.3 & 83.6 & 65.3 & 83.6 & 90.0 & 72.6 & 62.4 & 78.4 & 82.7 & 94.1 & 84.0 & 90.7 & 70.0 & 92.2 & 92.1 & 79.7 & 74.4 & 92.2 & 54.6 & 95.3 & 94.2 & 66.0 & \uline{80.9}\\
        SBI\cite{shiohara2022detecting} & 79.4&89.9&57.2&84.4&88.3&77.5&61.2&66.0 &72.2&79.9&64.4&84.8&68.8&79.8&82.3&50.9&63.9&88.0&45.1&89.2&88.5&61.4&73.8\\
        CORE\cite{ni2022core} & 80.9 & 79.6 & 56.5 & 81.7 & 88.6 & 79.7 & 54.6 & 75.6 & 67.5 & 82.5 & 79.7 & 82.4 & 57.9 & 83.2 & 83.6 & 73.8 & 66.2 & 91.7 & 53.2 & 87.6 & 91.0 & 51.9 &75.0\\ 
        SIA\cite{sun2022information} & 79.2 & 66.9 & 57.1 & 70.3 & 70.8 & 71.0 & 60.0 & 67.4 & 66.5 & 74.6 & 66.1 & 65.8 & 51.2 & 71.2 & 73.0 & 69.7 & 70.1 & 74.3 & 64.5 & 84.2 & 82.5 & 65.8 &69.2\\
        UCF\cite{yan2023ucf} & 83.7 & 80.6 & 55.4 & 82.1 & 85.0 & 78.5 & 56.6 & 74.8 & 71.2 & 87.7 & 78.9 & 85.9 & 58.0 & 85.5 & 86.0 & 76.5 & 67.6 & 88.8 & 64.2 & 88.3 & 88.7 & 49.6 &76.1\\
        IID\cite{huang2023implicit} & 80.8 & 82.8 & 60.0 & 80.3 & 84.5 & 76.0 & 58.6 & 73.4 & 74.2 & 89.4 & 77.3 & 86.9 & 65.4 & 87.2 & 88.1 & 71.8 & 66.0 & 85.6 & 49.1 & 89.9 & 86.3 & 57.3 &76.0\\
        LSDA\cite{yan2024transcending} & 77.6 & 81.3 & 60.8 & 82.1 & 81.1 & 66.4 & 50.7 & 74.4 & 65.6 & 82.9 & 69.9 & 80.2 & 64.3 & 79.4 & 80.3 & 65.9 & 69.9 & 81.8 & 58.5 & 82.4 & 88.9 & 55.5 &72.7\\
        CFM\cite{luo2023beyond} & 87.5 & 85.5 & 65.3 & 80.8 & 82.4 & 74.3 & 57.8 & 82.6 & 68.0 & 83.9 & 70.6 & 87.0 & 64.0 & 86.6 & 88.7 & 70.0 & 74.1 & 88.7 & 48.6 & 91.2 & 87.4 & 58.7 &76.5\\
        ProDet\cite{cheng2024can} & 92.6 & 90.3 & 56.2 & 87.2 & 89.9 & 90.1 & 52.1 & 80.7 & 50.6 & 69.2 & 75.4 & 72.8 & 56.1 & 67.6 & 70.6 & 62.8 & 68.8 & 88.0 & 69.8 & 83.1 & 91.9 & 43.6 &73.2\\
        ForAda\cite{cui2024forensics} & 95.7 & 90.1 & 71.7 & 89.5 & 92.7 & 89.5 & 75.3 & 92.4 & 63.1 & 67.6 & 76.0 & 72.7 & 56.9 & 73.3 & 70.1 & 71.3 & 59.8 & 89.1 & 57.8 & 76.6 & 77.2 & 51.1 &75.4\\
        Effort\cite{yan2024effort} & 93.8 & 90.7 &  72.1 & 87.0 & 90.2 & 84.7 & 72.0 & 96.4 & 79.4 & 88.1 & 91.5 & 88.1 & 61.5 & 90.0 & 88.5 & 85.0 & 81.8 & 96.8 & 83.2 & 90.2 & 95.8 & 62.2 &\textbf{85.0}\\
        \hline
        Average &80.9 &78.9 &59.0 &78.9 &82.0 &74.6 &{58.4} &74.0 &69.0 &80.9 & 73.9&79.1 &59.5 &79.2 &80.1 &69.9 &68.7 & 83.9&58.9 &{86.3} & {86.1} &58.6 &73.7 \\
    \hline 
    \end{tabular}}
\end{table*}

\smallskip
\noindent\textbf{Protocol \#1 (GF-eval).}
Under this protocol, all detection methods \textit{are trained only on Celeb-DF in the Face-swap scenario and tested on all other DeepFake methods across Face-swap, Face-reenactment, and Talking-face scenarios. }

Since several detection methods have not fully released their training codes, we select eight representative ones and retrain them under this protocol. 
As shown in Table~\ref{tab:cross_method} and Table~\ref{tab:cross_method_video}, these detection methods only achieve approximately $71.7\%$ and $72.1\%$ on average, respectively,  demonstrating their limitation on generalizability across different DeepFake methods. Even in the same scenario, the detection performance drops significantly across different types of DeepFake methods, underscoring sensitivity to method-specific artifacts. Note that Effort achieves the highest average frame-level AUC of $83.0\%$ and video-level AUC of $84.4\%$, showing a favorable intra-scenario performance. However, its performance degrades significantly when applied to other scenarios, demonstrating its limited cross-scenario detection ability. These findings emphasize the value of this evaluation protocol and offer instructive insights to enhance the generalizability of detection methods.  


\smallskip
\noindent\textbf{Protocol \#2 (GFQ-eval).}
In real-world social media platforms, video content often undergoes varying degrees of lossy compression during uploading and downloading. These compressions can degrade their visual quality, thus impairing the performance of DeepFake detectors. To simulate this scenario, we use the FFmpeg tool~\cite{tomar2006converting} to compress DeepFake videos with various factors. Specifically, \textit{we adopt the H.264 encoding standard and configure two compression levels: \textbf{c35} and \textbf{c45}, corresponding to medium and high compression strength, respectively. All other experimental configurations follow the GF-eval protocol.}

Table~\ref{tab:cross_quality_c35}, \ref{tab:cross_quality_c35_video}, \ref{tab:cross_quality_c45} and \ref{tab:cross_quality_c45_video} show the performance of existing detectors under two compression levels at both the frame- and video-level. The results show a clear degradation: Under c35 compression, frame-level performance drops by an average of $3.5\%$, and under the stronger c45 compression, the average drop increases to $4.4\%$, while at the video-level, the drops are $2.2\%$ and $7.5\%$, respectively. These findings confirm that heavier compression can more effectively obscure forgery traces, making detection more difficult. Using this protocol highlights the limited generalizability of existing detection methods when confronted with compression commonly found in real-world scenarios. 


\smallskip
\noindent\textbf{Protocol \#3 (GFD-eval).}
In addition to considering cross-quality in GFQ-eval, we also investigate the performance of detection methods in a cross-dataset scenario. This is both practical and challenging, as it reflects real-world applications where the training and testing data originate from different sources. Since all detectors provide weights trained FF++ (HQ), we adopt the common setting described in Sec.~\ref{sec:challenge}, where \textit{each detector is trained on the FF++ (HQ) dataset and tested on individual DeepFake methods within Celeb-DF++.} 

The frame-level and video-level AUC scores are reported in Table~\ref{tab:22-frame-level} and Table~\ref{tab:22-video-level}, respectively. Since FF++ (HQ) mainly contains Face-swap DeepFakes, all detectors perform better in the Face-swap scenario of Celeb-DF than in the Face-reenactment and Talking-face scenarios. However, compared to the results in GF-eval, the results of this protocol degrade notably due to the domain gap between FF++ and Celeb-DF++, highlighting the necessity of developing a generalizable DeepFake detector against data domain shifts.

\section{Conclusion}
In this paper, we introduce Celeb-DF++, a diverse and challenging large-scale DeepFake benchmark dedicated for \textit{generalizable forensics} in DeepFake detection. 
Built upon our earlier Celeb-DF dataset, Celeb-DF++ incorporates a wider range of recent DeepFake methods, spanning three commonly observed forgery scenarios: Face-swap (FS), Face-reenactment (FR), and Talking-face (TF). Each scenario contains a large collection of high-quality forged videos, generated using 8, 7, and 7 various DeepFake methods, respectively. Moreover, we describe three new evaluation protocols for measuring the generalizability of detection methods. Compared to existing benchmarks, Celeb-DF++ studies a boarder range of recent detection methods, covering 19 classic methods and 5 state-of-the-art methods introduced after 2024. Experimental results highlight that generalizable DeepFake detection remains a highly challenging task while demonstrate the difficulty of our new benchmark.

\bibliography{ref}
\bibliographystyle{IEEEtran}
\vfill

\end{document}